%% file: ba_map.tex
\documentclass[preprint,12pt]{elsarticle}
\input{preamble}

\begin{document}

\begin{frontmatter}
  \input{frontmatter}
\end{frontmatter}

\section{Introduction}\label{sec:intro}
\input{intro}

\section{Problem Formulation}\label{sec:problem}
\input{problem}

\section{Methodology}\label{sec:methodology}
\input{methodology}

\section{Numerical Experiments}\label{sec:experiments}
\input{experiments}

\section{Conclusions}\label{sec:conclusions}
\input{conclusions}

\section{Acknowledgments}
This research was supported by the U.S. Department of Energy (DOE) Advanced Scientific Computing (ASCR) program. Pacific Northwest National Laboratory is operated by Battelle for the DOE under Contract DE-AC05-76RL01830.

\bibliographystyle{elsarticle-num} 
\bibliography{ba_map}

\end{document}

%% file: preamble.tex
\usepackage[dvipsnames]{xcolor}
\usepackage{url} 
\usepackage{soul}
\usepackage{lineno}
\usepackage{inputenc}
\usepackage[export]{adjustbox}
\usepackage{amsmath, amsfonts, amssymb, amsthm, bm, mathtools}
\usepackage{algorithm, algpseudocode}
\usepackage{natbib}
\usepackage{tabularray}
\UseTblrLibrary{amsmath, siunitx}
\usepackage[export]{adjustbox}
\usepackage{caption}
\usepackage[subrefformat=parens]{subcaption}
\usepackage{hyperref}
\usepackage[capitalise]{cleveref}
\usepackage{xstring}
\usepackage{etoolbox}
\usepackage[mode=buildnew]{standalone}
\usepackage{tikz, pgfplots, pgfplotstable}
\usetikzlibrary{backgrounds, calc, external, matrix, positioning, tikzmark}
\usepgfplotslibrary{external, groupplots, statistics}

\pgfplotscreateplotcyclelist{pdfcolorlist}{%
    Maroon, dashed, mark=none\\%
    MidnightBlue, mark=none\\%
    SeaGreen, mark=none\\%
    Orange, only marks, mark=o, mark repeat=50, mark phase=25\\%
}

\pgfplotscreateplotcyclelist{scattercolorlist}{%
    ultra thin, only marks, mark options = {draw = MidnightBlue, fill = MidnightBlue, scale = 0.5} \\%
}

\pgfplotsset{
    compat = newest,
    every axis/.append style = {
        thick,
        line width = 1pt,
        tick label style = {
            font = \footnotesize,
        },
        x tick label style = {
            /pgf/number format/.cd,
            precision = 0,
            set thousands separator = {},
            /tikz/.cd
        },
        legend style = {
            font = \footnotesize,
        },
    },
    every axis legend/.append style = {
        line width = 1pt,
    },
    scatter plots/.style = {
        height = 0.18\textheight,
        width = 0.18\textheight,
        cycle list name = scattercolorlist,
    },
    pdf plots/.style = {
        height = 0.2\textheight,
        width = 0.36\textwidth,
        cycle list name = pdfcolorlist,
        y tick label style = {
            /pgf/number format/.cd,
            fixed,
            fixed zerofill,
            precision = 2,
            /tikz/.cd
        },
        legend columns=-1,		
        legend cell align=left,
        legend style={/tikz/every even column/.append style={column sep=1em}},
        legend entries={Actual, BA 2$\times$1D},
    },
    single plot/.style = {
        width = 0.85*\textwidth,
        height = 0.3*\textheight,
    },
    error plot/.style = {
        single plot,
        cycle list = {
            Maroon,every mark/.append style={fill=Maroon,solid},mark=*\\%
            MidnightBlue,every mark/.append style={fill=MidnightBlue,solid},mark=square*\\%
        },
        xlabel={Number of KL terms},
        ylabel={$\ell_2$ errors},
    },
    trend plot/.style = {
        all plot,
        cycle multi list={
            {mark=*}, {mark=square*}\nextlist
            MidnightBlue, {Maroon, draw=none}\nextlist
            only marks, mark=none
        },
        xlabel={Number of FV cells},
        ylabel={Execution time (s)},
        xtick=data,
        xticklabel={
            \pgfkeys{/pgf/fpu=true}
            \pgfmathparse{exp(\tick)}
            \pgfmathprintnumber[fixed relative, precision=4]{\pgfmathresult}
            \pgfkeys{/pgf/fpu=false}
        },
    },
}

\tikzset {
    every picture/.append style = {
        thick,
        every node/.style = {font = \footnotesize},
    },
}

\newcommand{\colorbar}[3]{
\tikzset{external/export next=false}
\begin{tikzpicture}[baseline={(0, -2.75em)}]
    \pgfplotscolorbardrawstandalone[
        parent axis height/.initial = 5em,
        point meta min=#1,
        point meta max=#2,
        colormap/jet,
        colorbar,
        colorbar style={
            width=0.4em,
            ytick=#3
    }]
\end{tikzpicture}
}

\newcommand*{\minmax}[4]{%
    \pgfmathsetmacro#3{-16383}%
    \pgfmathsetmacro#4{16383}%
    \pgfplotsforeachungrouped \i in {#2} {
        \pgfplotstableforeachcolumnelement{\i}\of{#1}\as\cellValue{%
            \ifx\cellValue\@empty\else
                \pgfmathsetmacro{#3}{max(#3,\cellValue)}%
                \pgfmathsetmacro{#4}{min(#4,\cellValue)}%
            \fi
        }
    }
}

\newcommand\setv[3]{\expandafter\xdef\csname #1_#2\endcsname{#3}}
\newcommand\getv[2]{\csname #1_#2\endcsname}

\newcommand*{\vr}{\rule[-1ex]{0.5pt}{1.5em}}
\newcommand*{\hr}{\rule[.5ex]{1.5em}{0.5pt}}
\DeclareMathOperator*{\argmin}{arg\,min}

\makeatletter
\renewcommand*\env@matrix[1][*\c@MaxMatrixCols c]{%
  \hskip -\arraycolsep
  \let\@ifnextchar\new@ifnextchar
  \array{#1}}
\makeatother

\adjustboxset{
    keepaspectratio,
    valign = M
}

\BeforeBeginEnvironment{equation}{\begin{linenomath}}
\AfterEndEnvironment{equation}{\end{linenomath}}
\BeforeBeginEnvironment{equation*}{\begin{linenomath}}
\AfterEndEnvironment{equation*}{\end{linenomath}}
\BeforeBeginEnvironment{gather}{\begin{linenomath}}
\AfterEndEnvironment{gather}{\end{linenomath}}
\BeforeBeginEnvironment{align}{\begin{linenomath}}
\AfterEndEnvironment{align}{\end{linenomath}}


%% file: frontmatter.tex
\title{
Conditional Korhunen-Lo\'{e}ve regression model with Basis Adaptation for high-dimensional problems: uncertainty quantification and inverse modeling
}
\author[1]{Yu-Hong Yeung}
\ead{yhy@illinois.edu}

\author[2]{Ramakrishna Tipireddy}
\ead{rtipireddy@paloaltonetworks.com}

\author[3]{David A. Barajas-Solano}
\ead{David.Barajas-Solano@pnnl.gov}

\author[1,3]{Alexandre M. Tartakovsky}
\ead{amt1998@illinois.edu}

\affiliation[1]{%
  organization={Department of Civil and Environmental Engineering, University of Illinois Urbana-Champaign},%
  city={Urbana},%
  postcode={61801},%
  state={IL},%
  country={USA}%
}

\affiliation[2]{
  organization={Palo Alto Networks},
  city={Santa Clara},
  postcode={95054},
  state={CA},
  country={USA}
}

\affiliation[3]{%
  organization={Physical and Computational Sciences Directorate, Pacific Northwest National Laboratory},%
  city={Richland},%
  postcode={99354},%
  state={WA},%
  country={USA}%
}

\begin{abstract}
  We propose a methodology for improving the accuracy of surrogate models of the observable response of physical systems as a function of the systems' spatially heterogeneous parameter fields with applications to uncertainty quantification and parameter estimation in high-dimensional problems. 
  Practitioners often formulate finite-dimensional representations of spatially heterogeneous parameter fields using truncated unconditional Karhunen-Lo\'{e}ve expansions (KLEs) for a certain choice of unconditional covariance kernel and construct surrogate models of the observable response with respect to the random variables in the KLE.
  When direct measurements of the parameter fields are available, we propose improving the accuracy of these surrogate models by representing the parameter fields via \emph{conditional} Karhunen-Lo\'{e}ve expansions (CKLEs).
  CKLEs are constructed by conditioning the covariance kernel of the unconditional expansion on the direct measurements via Gaussian process regression and then truncating the corresponding KLE.
  We apply the proposed methodology to constructing surrogate models via the Basis Adaptation (BA) method of the stationary hydraulic head response, measured at spatially discrete observation locations, of a groundwater flow model of the Hanford Site, as a function of the $1,000$-dimensional representation of the model's log-transmissivity field.
  We find that BA surrogate models of the hydraulic head based on CKLEs are more accurate than BA surrogate models based on unconditional expansions for forward uncertainty quantification tasks.
  Furthermore, we find that inverse estimates of the hydraulic transmissivity field computed using CKLE-based BA surrogate models are more accurate than those computed using unconditional BA surrogate models.
\end{abstract}

\date{January 2023}


%% file: intro.tex
We propose a novel surrogate modeling approach with application to uncertainty quantification and parameter estimation in high-dimensional problems.

Uncertainty quantification and parameter estimation for physical systems with unknown or uncertain spatially heterogeneous parameter fields are often very high-dimensional problems.
This is because a large number of degrees of freedom are necessary to fully represent the various scales of spatial heterogeneity that characterize these parameter fields.
High dimensionality makes uncertainty quantification and parameter estimation highly challenging problems due to the combination of two factors:
first, computational methods for these tasks generally require a number of forward solver queries that, with a few exceptions such as simple Monte Carlo sampling, scales with the system's number of degrees of freedom;
second, the boundary value problem (BVP) forward solvers for these systems are often computationally expensive.
This, consequently, makes the computational costs of these tasks unfeasible.

Recently, scientific machine learning methods have been proposed for uncertainty quantification and parameter estimation~\cite{yang2019highly,he2021physics,yeung-2022-wrr,yeung-2023-jcp,psaros-uq-2023,yang-bpinns-2021,raissi-pinns-2019,tartakovsky-pickle-2021}.
These methods aim to leverage the capabilities of machine learning methods to model high-dimensional data for solving scientific computing problems.
For high-dimensional problems with number of dimensions $O(10^3)$ and larger, state-of-the-art scientific machine learning methods require large amounts of data.
The physics-informed GAN model proposed in~\cite{yang2019highly} for uncertainty quantification in a groundwater flow model of the Hanford Site, a U.S. Department of Energy site located in Washington State, requires a large volume of training data.
Training and prediction for this physics-informed GAN model were performed on the Department of Energy Oak Ridge Leadership Computing Facility ``Summit'' supercomputer.
For parameter estimation in the same groundwater model of the Hanford Site, the physics-informed conditional Karhunen-Lo\`{e}ve method (PICKLE)~\cite{tartakovsky-pickle-2021} was shown to be more efficient than the traditional PDE-constrained maximum a posteriori (MAP) estimation method.
It was found that the computational cost of PICKLE scales linearly with the number of finite volume cells of the Hanford Site model, in comparison to cubic scaling for MAP.
Despite this advantage, PICKLE requires thousands of forward simulations for training~\cite{yeung-2022-wrr}.

A possible strategy for addressing the challenges posed by high dimensionality and the high cost of forward solver queries is the construction of low-dimensional surrogate models.
Let $\bm{\xi} \in \mathbb{R}^{N_{\xi}}$ denote the system's vector of degrees of freedom, and $\mathbf{u} \colon \mathbb{R}^{N_{\xi}} \mapsto \mathbb{R}^{N_u}$ the system's observation function, which maps the sytem's degrees of freedom to the system's observable response.
The map $\bm{\xi} \mapsto \mathbf{u}(\bm{\xi})$ is implicitly defined by the forward solver, and we assume that this map is injective.
A surrogate model is a function $\mathbf{f} \colon \mathbb{R}^{N_{\xi}} \mapsto \mathbb{R}^{N_u}$ that approximates the map $\bm{\xi} \mapsto \mathbf{u}(\bm{\xi})$ to some degree of accuracy and whose evaluation (i) does not require the use of the forward solver and (ii) is significantly faster than the forward solver.
Surrogate models are generally constructed in a data-driven, supervised manner via multivariate regression from a dataset of pairs $\{ \bm{\xi}, \mathbf{u}(\bm{\xi}) \}$.

A common method for constructing surrogate models is generalized Polynomial Chaos (gPC)~\cite{zhou-surrogate-2020,novak-polynomial-2018}.
However, the number of coefficients in the PC expansion (and the required number of forward solutions for estimating these coefficients) exponentially increases with $N_\xi$ in a manifestation of the phenomenon known as the ``curse of dimensionality''~\cite{tipireddy-2014-jcp,li-inverse-2016,peng-2014-jcp}.
Generating the training dataset and training the surrogate models add to the time-to-solution, so it is important to ensure that these additional costs are feasible.
Therefore, gPC-based surrogates are not feasible for high-dimensional problems unless the dimension of the input space is reduced.

For many high-dimensional systems (i.e., systems with large $N_\xi$), the observation function $\mathbf{u}(\bm{\xi})$ has a low-dimensional structure, which can be exploited to reduce the cost of training surrogate models, both in terms of training time and in the required amount of training data~\cite{zhou-surrogate-2020,hou-2022-dse}.
For such systems, the observation function can be approximately characterized by a set of $r \ll N_{\xi}$ ``effective'' coordinates $\bm{\eta}(\bm{\xi})$ to some degree of accuracy (this is also one of the reasons why inverse problems for such systems are ill-posed: due to this low-dimensional structure, measurements of $\mathbf{u}(\bm{\xi}^{*})$ are not sufficient to fully identify the data-generating $\bm{\xi}^{*}$).
We can then formulate a low-dimensional surrogate for this observation function that consists of the composition of $\bm{\eta}(\bm{\xi})$ and the lower dimensional regression function $\mathbf{f}(\bm{\eta})$.
Identifying the transformations $\bm{\xi} \mapsto \bm{\eta}(\bm{\xi})$ and $\bm{\eta} \mapsto \mathbf{f}(\bm{\eta})$ are the main tasks of low-dimensional surrogate modeling.

Various methods have been proposed for identifying the transformation to effective coordinates for a given observation function.
These methods can be classified into methods for linear transformations and methods for nonlinear transformations.
Methods for linear transformations such as active subspaces (AS)~\cite{constantine-active-2015}, basis adaptation (BA)~\cite{tipireddy-2018-juq,tipireddy-2017-jcp,tipireddy-2014-jcp,zeng-2021-cmame}, and sliced inverse regression (SIR)~\cite{li-inverse-2016,li-sir-1991}, among others, assume that the variation of the observable function is concentrated over an $r$-dimensional linear subspace of parameter space and formulate the transformation as a linear transformation $\bm{\eta} = \mathbf{A} \bm{\xi}$ for a $\mathbf{A} \in \mathbb{R}^{r \times N_{\xi}}$ to be identified.
AS requires access to the gradients $\partial \mathbf{u} / \partial \bm{\xi}$, while BA and SIR only require zeroth-order information.
While having access to gradient information is useful, methods that only require zeroth-order information are valuable given that many legacy forward solvers do not have the capabilities to output gradient information.
Therefore, in this work we will use the BA method for dimension reduction.
Nonlinear methods employ either kernelization-based~\cite{yeh-nonlinear-2009,bach-kica-2002} or deep learning-based algorithms~\cite{bigoni-nonlinear-2022,tripathy-deep-2018,bridges-active-2019,zhang-learning-2019} to identify nonlinear transformations to effective coordinates, but will not be considered in this work.

Direct, spatially sparse measurements of the parameter field may be available for certain problems.
For these problems, in this work we propose to improve the accuracy of BA-based low-dimensional surrogate models of observable functions by constructing a finite-dimensional representation of the system's parameter fields conditioned on the direct measurements.
In practice, finite-dimensional representations of heterogeneous fields are constructed via Gaussian process regression (Kriging)-based methods such as ``pilot points''~\cite{alcolea-pilot-2006,certes-application-1991} and Karhunen-Lo\'{e}ve expansions (KLEs)~\cite{huang-convergence-2001}.
Recent work has proposed to represent parameter fields via ``conditional Karhunen-Lo\'{e}ve expansions'' (CKLEs) constructed by conditioning GPR models of the parameter fields on the direct measurements and then truncating the conditional GPR model's KLE~\cite{tipireddy-2020-jcp}.
For uncertainty quantification tasks, it has been found that CKLEs result in significant reduction of uncertainty and more accurate estimates of statistics of quantities of interest for the same number of degrees of freedom when compared to unconditional KLEs.
Furthermore, CKLEs have been employed for representing unknown parameter fields in PDE-constrained parameter estimation, resulting in reduced computational cost compared to grid-based parameterizations~\cite{yeung-2023-jcp}.

We demonstrate the efficiency of using a combination of CKLE and BA-based low-dimensional surrogate models for a 1000-dimensional problem of estimating the hydraulic head as a function of the transmissivity field in a two-dimensional groundwater model of the Hanford Site.
For uncertainty quantification tasks, we find that the accuracy of the proposed surrogate models for modeling the conditional response is higher than the accuracy of the surrogate models for modeling the unconditional response.
For parameter estimation tasks, we find that the proposed surrogate models lead to more accurate estimates of the unknown transmissivity field compared to surrogate models of the unconditional response.

This manuscript is structured as follows: In~\cref{sec:problem}, we formulate the uncertainty quantification and parameter estimation problems for physical systems modeled using BVPs with parameter fields represented via KLEs.
In~\cref{sec:methodology}, we present the CKLE construction and describe algorithms for constructing low-dimensional BA-based surrogate models with the CKLE coefficients as the input.
The application of the proposed algorithms to the Hanford Site groundwater model is described~\cref{sec:experiments}.
Finally, we present conclusions and possible future research directions in~\cref{sec:conclusions}.


%% file: problem.tex
We consider a physical system in the simulation domain $D \subset \mathbb{R}^{d}$, $d \in [1, 3]$, governed by the partial differential boundary value problem (BVP)
\begin{equation}
  \label{eq:bvp}
  \mathcal{L}(u(\cdot), y(\cdot)) = 0,
\end{equation}
where $\mathcal{L}$ denotes the governing equation and boundary conditions, $y \in L^{2}(D)$ is the system's spatially heterogeneous parameter field, $u \in \mathcal{U}$ is the system's state, and $\mathcal{U}$ is an appropriately chosen (depending on the problem and the choice of solution scheme for~\Cref{eq:bvp}) function space of solutions to the BVP.
This system is observed via a measurement operator $\mathbf{h}_u \colon \mathcal{U} \times L^{2}(D) \to \mathbb{R}^{N_u}$, where $N_u$ is the number of scalar observables or quantities of interest.

We assume that the BVP is well posed and that the initial and boundary conditions are known so that the BVP implicitly defines the injective solution operator $u = \mathcal{G}(y)$, $\mathcal{G} \colon L^{2}(D) \mapsto \mathcal{U}$.
We then define the ``observation function'' of the system as the function
\begin{equation}
  \label{eq:observable-response}
  \mathbf{g}(y) \coloneqq \mathbf{h}_u(\mathcal{G}(y), y).
\end{equation}

We are interested in the problem of constructing surrogate models for $\mathbf{g}(\cdot)$, to be used for solving uncertainty quantification and model inversion problems.
Our starting point is the finite-dimensional representation of $y$ using a KLE; namely, for a certain choice of mean function $m \colon D \mapsto \mathbb{R}$ and continuous, symmetric, and positive definite kernel $C \colon D \times D \mapsto \mathbb{R}$, we can approximate $y$ via the KLE expansion truncated to $N_{\zeta}$ terms
\begin{equation}
  \label{eq:kle}
  y(x) \approx \tilde{y}(x; \bm{\zeta}) \coloneqq m(x) + \bm{\phi}^{\top}(x) \Lambda^{1/2} \bm{\zeta},
\end{equation}
where $\bm{\zeta} \in \mathbb{R}^{N_{\zeta}}$ is the vector of KLE coefficients and $\Lambda \coloneqq \operatorname{diag}(\lambda_1, \ldots, \lambda_{N_{\zeta}})$ and $\bm{\phi}(x) \coloneqq [\phi_1(x), \dots, \phi_{N_{\zeta}}(x)]$ are the diagonal matrix of eigenvalues and the vector of eigenfunctions, in the sense of Mercer's theorem, of the kernel $C$; that is, $\{\lambda_i, \phi_i(x) \}^{N_{\zeta}}_{i = 1}$ are found by solving the eigenproblem
\begin{equation}
  \label{eq:eigenproblem}
  \int_D C(x, y) \phi(y) \, \mathrm{d} y = \lambda \phi(x).
\end{equation}
We can write the observation function in terms of the KLE coefficients by substituting~\cref{eq:kle} into~\cref{eq:observable-response}, that is, 
\begin{equation}
  \label{eq:observable-response-kle}
  \tilde{\mathbf{g}}(\bm{\zeta}) \coloneqq \mathbf{g}(\tilde{y}(\cdot; \bm{\zeta})).
\end{equation}

The uncertainty quantification problem consists of approximating the distribution of the observation function when the parameter field is a random field.
This problem is tackled within this framework as follows: we represent the square-integrable random field $y$ defined over the probability triple $(\Omega, \mathcal{F}, \mathbb{P})$, with mean $m(x)$ and covariance $C(x, y)$,  using the KLE~\labelcref{eq:kle} so that the KLE coefficients $\bm{\zeta}$ are random with some distribution $p(\bm{\zeta})$.
We then aim to estimate the probability distribution function (PDF) of $\tilde{\mathbf{g}}(\bm{\zeta})$, given by
\begin{equation*}
  p(\mathbf{u}) \coloneqq \int \delta \left ( \mathbf{u} - \tilde{\mathbf{g}}(\bm{\zeta}) \right ) \, p(\bm{\zeta}) \, \mathrm{d} \bm{\zeta}
\end{equation*}

The model inversion problem consists of estimating a reference field $y_{\mathrm{ref}}$ from measurements of the observable response and direct measurements of $y_{\mathrm{ref}}$.
The direct measurements are taken at the set of $N_{y}$ observation locations $X \coloneqq \{ x_{i} \in D \}^{N_{y}}_{i = 1}$.
In the sequel, we employ the notation $f(X)$ for a function $f \colon D \mapsto \mathbb{R}$ to denote the $N_y$-dimensional column vector $[f(x_i), \dots, f(x_{N_y})]^{\top}$.
Similarly, for the kernel function $C \colon D \times D \mapsto \mathbb{R}$, $C(x, X)$ denotes the row vector $[C(x, x_i), \dots, C(x, x_{N_y})]$ with $C(X, x) \equiv C(x, X)^{\top}$, and $C(X, X)$ denotes the matrix $[C(x_i, x_j)]_{ij}$.
We assume that the measurements are of the form
\begin{equation}
  \label{eq:measurement-operators}
  \hat{\mathbf{u}} \coloneqq \mathbf{g}(y_{\mathrm{ref}}) + \bm{\epsilon}_u, \quad \hat{\mathbf{y}} \coloneqq y_{\mathrm{ref}}(X) + \bm{\epsilon}_{y},
\end{equation}
where $\bm{\epsilon}_u$ and $\bm{\epsilon}_y$ are additive measurement errors with covariances $\sigma^2_u I$ and $\sigma^2_y I$, respectively.
We then formulate the model inversion problem in terms of the KLE~\labelcref{eq:kle} as the minimization problem
\begin{equation}
  \label{eq:inverse-problem}
  \min_{\bm{\zeta}} \, \frac{1}{2 \sigma^2_u} \| \hat{\mathbf{u}} - \tilde{\mathbf{g}}(\bm{\zeta}) \|^2_2 + \frac{1}{2 \sigma^2_y} \| \hat{\mathbf{y}} - \tilde{y}(X; \bm{\zeta}) \|^2_2 + \frac{\gamma}{2} \rho(\bm{\zeta}),
\end{equation}
where we have introduced $\rho \colon \mathbb{R}^{N_{\zeta}} \mapsto \mathbb{R}$ and the weight $\gamma > 0$ to regularize the inverse problem.


%% file: methodology.tex
\subsection{Representing \texorpdfstring{$y$}{y} using CKLEs}

As in~\cite{yeung-2022-wrr, yeung-2023-jcp}, we propose incorporating the direct measurements $\hat{\mathbf{y}}$ into the finite-dimensional representation of $y$.
Specifically, we propose approximating $y$ using a truncated \emph{conditional} KLE, defined as the KLE with mean $\bar{y}^c$ and kernel $C^c$ given by the Gaussian process regression (Kriging) equations
\begin{align*}
  \bar{y}^c(x) &= m(x) + C(x, X) \left [ C(X, X) + \sigma^2_y I \right ]^{-1} \left [ \hat{\mathbf{y}} - m(X) \right ],\\
  C^c(x, y) &= C(x, y) - C(x, X) \left [ C(X, X) + \sigma^2_y I \right ]^{-1} C(X, y).
\end{align*}
The CKLE truncated to $N_{\xi}$ terms reads
\begin{equation}
    \label{eq:ckle-y}
    y(x) \approx \tilde{y}^c(x; \bm{\xi}) \coloneqq \bar{y}^c(x) + (\bm{\phi}^c)^{\top}(x) (\Lambda^c)^{1/2} \bm{\xi}
\end{equation}
where $\boldsymbol{\xi} \in \mathbb{R}^{N_{\xi}}$ is the vector of CKLE coefficients, $\bm{\phi}^c \coloneqq [\phi^c_1(x), \dots, \phi^c_{N_{\xi}}(x)]$ is the vector of CKLE eigenfunctions, and $\Lambda^c = \operatorname{diag}(\lambda^c_1, \dots, \lambda^c_{N_{\xi}})$ is the diagonal matrix of CKLE eigenvalues.
The eigenpairs $\{ \lambda^c_i, \phi^c_i(x) \}$ are defined as the solution to the eigenproblem~\cref{eq:eigenproblem} by substituting $C^c$ for $C$.

We can write the observable response function in terms of the CKLE coefficients as
\begin{equation}
  \label{eq:observable-response-ckle}
  \tilde{\mathbf{g}}^c(\bm{\xi}) \coloneqq \mathbf{g}(\tilde{y}^c(\cdot; \bm{\xi})).
\end{equation}
By construction, $\hat{\mathbf{y}} - \tilde{y}^c(X; \bm{\xi})$ is $O(\sigma_y)$ for a given $\bm{\xi}$, which allows us to drop the second term in~\cref{eq:inverse-problem}; therefore, we can reformulate the inverse problem as
\begin{equation}
  \label{eq:inverse-problem_ckle}
  \min_{\bm{\xi}} \, \frac{1}{2 \sigma^2_u} \| \hat{\mathbf{u}} - \tilde{\mathbf{g}}^c(\bm{\xi}) \|^2_2+ \frac{\gamma}{2} R(\bm{\xi}),
\end{equation}
for a regularization function $R \colon \mathbb{R}^{N_\xi} \mapsto \mathbb{R}$.

\subsection{Surrogate modeling via BA}\label{sec:ba}

We formulate an element-wise ridge function approximation to the observation function as a low-dimensional surrogate model.
For the $i$th observable $\tilde{g}^c_i(\bm{\xi})$, this construction assumes that the variation of $\tilde{g}^c_i$ is concentrated over an $r^{(i)}$-dimensional linear subspace of $\mathbb{R}^{N_{\xi}}$, $V^{(i)}$, which we refer to as the ``latent space'', and that the variation of $\tilde{g}^c_i$ along $( V^{(i)} )^{\perp}$ can be disregarded, inducing a small approximation error.
Furthermore, we assume that $r^{(i)} \ll N_{\xi}$.
Specifically, we construct surrogate models of the form
\begin{equation}
  \label{eq:ridge-function-elementwise}
  \tilde{g}^c_i(\bm{\xi}) \approx f^{(i)}( \mathbf{A}^{(i)} \bm{\xi}),
\end{equation}
where $\mathbf{A}^{(i)}$ is a $\mathbb{R}^{r^{(i)} \times N_{\xi}}$ orthogonal rotation matrix satisfying $\mathbf{A}^{(i)} (\mathbf{A}^{(i)})^{\top} = \mathbf{I}_{r^{(i)}}$, whose row vectors form an orthogonal basis for the latent space $V^{(i)}$, and $f^{(i)} \colon \mathbb{R}^{r^{(i)}} \mapsto \mathbb{R}$ is a regression function.
The rotation matrix $\mathbf{A}^{(i)}$ projects $\bm{\xi}$ onto $V^{(i)}$, and we introduce the variable $\bm{\eta}^{(i)} \coloneqq \mathbf{A}^{(i)} \bm{\xi}$ over $V^{(i)}$.

We motivate the use of low-dimensional surrogate models for scalar observables by the following argument: If $\tilde{g}^c(\bm{\xi})$ is a linear scalar function $\tilde{g}^c(\bm{\xi}) = b + \mathbf{a}^{\top} \bm{\xi}$, its variation is restricted to the one-dimensional subdomain $\operatorname{span} \{ \mathbf{a} \}$.
This argument does not extend in general to nonlinear scalar observables, but as we find in~\cref{sec:experiments}, it is a good starting point for building surrogate models for the observables of the systems considered in this work.

We employ a data-driven approach to selecting $\mathbf{A}^{(i)}$ and $f^{(i)}(\cdot)$, $i \in [1, N_u]$.
For this purpose, we synthetically generate a dataset of $(\bm{\xi}, \tilde{\mathbf{g}}^c(\bm{\xi}))$ pairs as follows:
first, we select a data-generating distribution $p(\bm{\xi})$; then, we draw $q$ realizations $\bm{\xi}^{(k)} \sim p(\bm{\xi})$, $k \in [1, q]$ to generate the input dataset; finally we query the observation function for each $\bm{\xi}^{(k)}$, $\mathbf{u}^{(k)} \coloneqq \tilde{\mathbf{g}}^c(\bm{\xi}^{(k)})$, to generate the output dataset.
We organize the input and output data as the matrices
\begin{equation}
  \label{eq:data}
  \mathbf{U} =
  \begin{bmatrix}
    \vr & & \vr \\
    \tilde{\mathbf{g}}^c(\bm{\xi}^{(1)}) & \cdots & \tilde{\mathbf{g}}^c(\bm{\xi}^{(q)})\\
    \vr & & \vr    
  \end{bmatrix},
  \quad
  \bm{\Xi} =
  \begin{bmatrix}
    \vr & & \vr \\
    \bm{\xi}^{(1)} & \cdots & \bm{\xi}^{(q)}\\
    \vr & & \vr    
  \end{bmatrix}.
\end{equation}

A common application is when the indirect observables correspond to the field $u$ measured at observation locations $X_{u}$, that is, $\mathbf{h}_u(u, y) = u(X_u)$ and correspondingly $\mathbf{g}(y) = \mathcal{G}(y)(X_u)$.
In this case, to assemble the dataset we only need to evaluate the solution $u$ of the BVP~\cref{eq:bvp} given a certain field $\tilde{y}^c(\bm{\xi}^{*})$ at the locations $X_u$.
An acceleration technique can be used to evaluate $u$ at only those needed locations by exploiting the sparsity structure of the stiffness matrix of the BVP.
Details can be found in~\cite{yeung-2016-toc,yeung-2023-jcp}.

Because the surrogate model~\labelcref{eq:ridge-function-elementwise} is constructed element-wise, in the sequel we will omit the index $i$ indicating the $ith$ component $\tilde{g}^c_i$ of $\tilde{\mathbf{g}}^c$.
We first present an element-wise one-dimensional ($r = 1$) surrogate model (we find empirically that these models are accurate for the stationary scalar observables we consider in~\cref{sec:experiments}) and later discuss surrogate models with $r > 1$.
For $r = 1$, $\mathbf{A} \bm{\xi}$ reduces to $\hat{\mathbf{a}}^{\top} \bm{\xi}$.
We refer to $\hat{\mathbf{a}}$ as the ``dominant direction'', which satisfies $\hat{\mathbf{a}}^{\top} \hat{\mathbf{a}} = 1$.
The surrogate model then takes the form $f(\eta \coloneqq \hat{\mathbf{a}}^{\top} \bm{\xi})$.
We describe the construction of the BA surrogate model as follows: 
In~\cref{sec:dominant-direction}, we describe the data-based approach to identifying $\hat{\mathbf{a}}$ from the dataset defined by~\cref{eq:data}.
In~\cref{sec:pce_construction}, we present the polynomial chaos expansion (PCE) approach to identifying the nonlinear regressor $f(\eta)$.
Finally, in~\cref{sec:high-dimension-rotation} we discuss possible approaches for constructing the BA rotation matrix in the case $r > 1$.

\subsubsection{Dominant direction}\label{sec:dominant-direction}

Let $u(\bm{\xi}) \coloneqq \tilde{g}^c(\bm{\xi})$, and $\mathbf{u} = [u(\bm{\xi}^{(1)}), \dots, u(\bm{\xi}^{(q)})]^{\top}$ be the transposed $i$th row of the data matrix $\mathbf{U}$.
We introduce the normalized data vector $\hat{\mathbf{u}} \coloneqq (\mathbf{u} - \bar{u} \mathbf{1}) / \sigma_{\mathbf{u}}$, where $\bar{u}$ and $\sigma_{\mathbf{u}}$ are the ensemble mean and standard deviation of $\mathbf{u}$, and $\mathbf{1}$ is a $q \times 1$ vector of $1$s.
We find the dominant direction $\hat{\mathbf{a}}$ by modeling the map $\bm{\xi} \mapsto u(\bm{\xi})$ as an affine transformation.
Specifically, we model the transformation from the dataset $\bm{\Xi}$ to the normalized vector $\hat{\mathbf{u}}$ as
\begin{equation}
  \label{eq:dominant-direction-linear-model}
  \begin{aligned}
    \hat{\mathbf{u}} \approx \bm{\Xi}^{\top} \mathbf{a} + b \mathbf{1} &= \left[\boldsymbol{\Xi}^\top| \mathbf{1} \right] \left[\dfrac{\mathbf{a}}{b}\right] \\ %
    & = 
      \begin{bmatrix}[c|c]
        \hr \: {\boldsymbol{\xi}^{(1)}}^\top \: \hr & 1 \\
        \vdots & \vdots\\
        \hr \: {\boldsymbol{\xi}^{(q)}}^\top \: \hr & 1
      \end{bmatrix}
      \begin{+bmatrix}[hline{4} = {solid}]
        a^{(1)} \\
        \vdots\\
        a^{(m)} \\
        b
      \end{+bmatrix},
  \end{aligned}
\end{equation}
where $b$ is the bias and $\mathbf{a}$ is the vector of coefficients of the affine transformation.
We find $b$ and $\mathbf{a}$ by solving the basis pursuit denoising problem
\begin{equation}
  \label{eq:bpdn}
  \mathbf{a}^*, b^{*} = \argmin_{\mathbf{a}, b} \left\| \left[\dfrac{\mathbf{a}}{b}\right] \right\|_1 \quad \text{subject to} \quad \left\| \hat{\mathbf{u}} - \left[\boldsymbol{\Xi}^\top| \mathbf{1} \right] \left[\dfrac{\mathbf{a}}{b}\right] \right\|_2 \leq \epsilon,
\end{equation}
for some tolerance $\epsilon$.
Note that we only employ~\cref{eq:bpdn,eq:dominant-direction-linear-model} to estimate the dominant direction and that~\Cref{eq:dominant-direction-linear-model} is not used for predicting the system's response.
Once we have estimated $\mathbf{a}^{*}$, we discard the estimated bias $b^{*}$.
We obtain the dominant direction by normalizing $\mathbf{a}^*$, that is, $\hat{\mathbf{a}} = \mathbf{a}^* / \|\mathbf{a}^*\|_2$.

\subsubsection{PCE-based regression}\label{sec:pce_construction}

In this section, we discuss the construction of the nonlinear regressor using PCEs for the general case $r \geq 1$.
We represent the nonlinear regressor $f(\bm{\eta} \coloneqq \hat{\mathbf{a}}^{\top} \bm{\xi})$ using the truncated polynomial chaos expansion
\begin{equation}
    \label{eq:pce}
    f(\bm{\eta}) = \sum\nolimits_{\bm{\alpha} \in \mathcal{J}_{r, p}} c_{\bm{\alpha}} H_{\bm{\alpha}}(\bm{\xi}),
\end{equation}
where $\bm{\alpha} = (\alpha_1, \dots, \alpha_r) \in \mathcal{J}_{r, p} \subseteq (\mathbb{N}_0)^r$ is a $r$-dimensional multi-index, $\mathcal{J}_{r, p}$ is the subset of indices of maximum length $p$ (that is, $| \bm{\alpha} | \leq p$), $H_{\bm{\alpha}}$ is the normalized multivariate Hermite polynomial given by $H_{\bm{\alpha}} = h_{\bm{\alpha}} / \| h_{\bm{\alpha}} \|_{L^2}$, and $\{ h_{\bm{\alpha}} \}$ is the set of $r$-dimensional multivariate probabilist's Hermite polynomials.

If $f(\bm{\eta})$ is known, we can compute the coefficients $c_{\bm{\alpha}}$ in~\cref{eq:pce} by using the orthogonality property of the Hermite polynomials to express $c_{\bm{\alpha}}$ as the integral
\begin{equation}
  \label{eq:sparse_grid_prev}
  c_{\bm{\alpha}} = \frac{1}{(2 \pi)^{r / 2}} \int f(\bm{\eta}) H_{\bm{\alpha}}(\bm{\eta}) \exp \left ( -\frac{1}{2} \bm{\eta}^{\top} \bm{\eta} \right ) \, \mathrm{d} \bm{\eta}, \quad \bm{\alpha} \in \mathcal{J}_{r, p},
\end{equation}
which we approximate using a Smolyak-Gauss-Hermite quadrature rule, resulting in the approximation
\begin{equation}
  \label{eq:sparse_grid-prev}
  c_{\bm{\alpha}} \approx \sum^p_{i = 1} w_i f(\bm{\eta}^{(i)}) H_{\bm{\alpha}}( \bm{\eta}^{(i)}), \quad \bm{\alpha} \in \mathcal{J}_{r, p},
\end{equation}
where $\{ w_i, \bm{\eta}^{(i)} \}^p_{i = 1}$ is the set of Smolyak-Gauss-Hermite quadrature weights and quadrature points.
Unfortunately, $f(\bm{\eta})$ itself is not known; rather, we can query $\tilde{g}^c$ for a given $\bm{\xi}^{*}$, which corresponds to the query $f(\bm{\eta}^{*} = \mathbf{A} \bm{\xi}^{*})$.
Therefore, we need to translate the quadrature points $\bm{\eta}^{(i)}$ into points in $\bm{\xi}$-space to query the observable function.

For this, we note that every $\bm{\eta}^{*}$ is the image of at least one $\bm{\xi}$ because the projection $\mathbf{A} \bm{\xi}$ is surjective.
Consequently, we can find a preimage of any $\bm{\eta}^{*}$.
One such preimage is the least-square preimage $\boldsymbol{\xi}^{*} = \mathbf{A}^{+} \bm{\eta}^{*} \equiv \mathbf{A}^{\top} \bm{\eta}^{*}$, which is the least-square vector satisfying $\bm{\eta}^{*} = \mathbf{A} \bm{\xi}^{*}$.
Therefore, we replace $f(\bm{\eta}^{(i)})$ with $\tilde{g}^c(\mathbf{A}^{\top} \bm{\eta}^{(i)})$, obtaining the expression
\begin{equation}
  \label{eq:sparse_grid}
  c_{\bm{\alpha}} \approx \sum_{i=1}^{p} w_i \tilde{g}^c(\mathbf{A}^{\top} \bm{\eta}^{(i)}) H_{\bm{\alpha}}(\bm{\eta}^{(i)}).
\end{equation}

\subsubsection{Higher-dimensional case}\label{sec:high-dimension-rotation}

As stated before, the variation of nonlinear observables may be concentrated over subspaces with $r > 1$.
The BA formulation presented in~\cref{sec:pce_construction,sec:dominant-direction} is not capable of identifying a rotation matrix $\mathbf{A}$ with $r > 1$.
An extension to BA was recently proposed\cite{zeng-2021-cmame} capable of identifying dominant directions of variation for $r > 1$, but this extension requires computing integrals over $\mathbb{R}^{N_{\xi}}$ of products of multivariate Hermite polynomials, which can be prohibitively expensive for large $N_{\xi}$, such as in the numerical experiments presented in~\cref{sec:experiments}.
Therefore, we will not consider this approach in the present work.

Let $\hat{\mathbf{a}}_k$ denote the $k$th row of $\mathbf{A}$, transposed.
The formulation of~\cref{sec:pce_construction,sec:dominant-direction} can be extended to $r > 1$ by assuming that the variation of $\tilde{g}^{c}$ along $\operatorname{span} \{ \hat{\mathbf{a}}_1 \}$ is much larger than the variation along $\operatorname{span} \{ \hat{\mathbf{a}}_2, \dots, \hat{\mathbf{a}}_r\}$, that the variation along $\operatorname{span} \{ \hat{\mathbf{a}}_1, \hat{\mathbf{a}}_2 \}$ is much larger than the variation along $\operatorname{span} \{ \hat{\mathbf{a}}_3, \dots, \hat{\mathbf{a}}_r\}$, and so forth.
Under this assumption, we propose identifying the $(k + 1)$th dominant direction by applying the procedure of~\cref{sec:dominant-direction} to the difference between the data vector $\mathbf{u}$ and the predictions of the $k$-dimensional BA-PCE model.

Specifically, assume we have identified the first $k$ dominant directions.
Let $\mathbf{A}_k = [\hat{\mathbf{a}}_1, \dots, \hat{\mathbf{a}}_k]^{\top}$, and let $f_k(\mathbf{A}_k \bm{\xi})$ denote the $k$-dimensional BA-PCE model constructed for $\mathbf{A}_{k}$ using the procedure of~\cref{sec:pce_construction}.
We define the $(k + 1)$ residual data as
\begin{equation}
  \label{eq:u_res}
  \mathbf{u}_{k + 1} = \mathbf{u} - f_k(\mathbf{A}_k \bm{\Xi}),
\end{equation}
which is then normalized, resulting in the normalized vector of residuals $\hat{\mathbf{u}}_{k + 1}$.
We identify $\hat{\mathbf{a}}_{k + 1}$ by solving~\cref{eq:bpdn} with $\hat{\mathbf{u}}_{k + 1}$ in place of $\hat{\mathbf{u}}$ and
\begin{equation}
    \label{eq:project_xi}
    \bm{\Xi}_{k + 1} = \left(\mathbf{I} - \mathbf{A}_k \mathbf{A}_k^\top\right) \bm{\Xi}_k
\end{equation}
in place of $\bm{\Xi}$.
We employ $\bm{\Xi}_{k + 1}$ to ensure that $\hat{\mathbf{a}}_{k + 1}$ is orthogonal to $\hat{\mathbf{a}}_1, \dots, \hat{\mathbf{a}}_k$.
Once we have obtained $\hat{\mathbf{a}}_{k + 1}$, we add it to the rotation matrix, and construct the $(k + 1)$-dimensional BA-PCE model $f_{k + 1}$ using the procedure of~\cref{sec:pce_construction}.
This procedure is repeated until all desired $K \leq r$ dominant directions are identified.
We refer to the resulting surrogate model as the ``BA-$K$D'' surrogate.

An alternative approach is to compose $f_k$ as the sum of one-dimensional functions $q_i(\eta_i)$ each given by a PCE, that is,
\begin{equation*}
  f_k(\bm{\eta}) = \sum\nolimits_i q_i(\eta_i \coloneqq \hat{\mathbf{a}}_i \bm{\eta_i}).
\end{equation*}
For this construction we proceed as above, except that once we find $\hat{\mathbf{a}}_{k + 1}$ we estimate the PCE coefficients of $q_{k + 1}$ as in~\cref{eq:sparse_grid-prev,eq:sparse_grid} by projecting $\tilde{g}^c(\hat{\mathbf{a}}_{k + 1} \eta^{(i)}_{k + 1}) - f_k(0)$ onto a one-dimensional Hermite polynomial basis.
We refer to the resulting surrogate model as the ``BA-$K \times$1D'' surrogate.
The complete algorithm for computing the BA-$K$D and BA-$K \times$1D surrogates is summarized in~\cref{alg:ba}.

\begin{algorithm}
    \caption{Basis Adaptation-based low-dimensional surrogate modeling}\label{alg:ba}
    \begin{algorithmic}[1]
      \Require Data $\{\mathbf{u}, \bm{\Xi} \}$, number of final dimensions $K$
      \State $\bm{\Xi} \to \bm{\Xi}_0$, $0 \to \mathbf{A}_0$
      \For {$k = 1, \dots, K$}
        \State Compute the residual data $\mathbf{u}_k$ using~\cref{eq:u_res}
        \State Compute $\hat{\mathbf{u}}_k = (\mathbf{u}_k - \hat{u}_k \mathbf{1}) / \sigma_{\mathbf{u}_k}$
        \State Project $\bm{\Xi}_k$ onto $(\operatorname{span} \{ \hat{\mathbf{a}}_1, \dots, \hat{\mathbf{a}}_{k - 1}\})^{\perp}$ using~\cref{eq:project_xi}
        \State Solve~\cref{eq:bpdn} for $\hat{\mathbf{u}}_k$ and $\bm{\Xi}_k$ and normalize it to obtain $\hat{\mathbf{a}}_k$
        \State $[\mathbf{A}^{\top}_{k - 1}, \hat{\mathbf{a}}_k] \to \mathbf{A}^{\top}_k$
        \State Construct $f_k(\cdot)$ using the procedure described in~\cref{sec:pce_construction}.
      \EndFor
      \State \Return $\mathbf{A}_K$ and $f_k(\cdot)$
    \end{algorithmic}
\end{algorithm}


%% file: experiments.tex
\subsection{Case Study}

We consider two-dimensional saturated flow in a heterogeneous porous medium over the Hanford Site domain $D \subset \mathbb{R}^2$.
Let $u \colon D \mapsto \mathbb{R}$ denote the hydraulic head field, and $T \colon D \mapsto \mathbb{R}^+$ denote the domain's transmissivity field.
The head field is governed by the BVP
\begin{align}
    \label{eq:pde}
    \nabla \cdot \left [ T(x) \nabla u(x) \right ] & = 0, && x \in D,\\
    \label{eq:pde-flux-bc}
    T(x) \nabla u(x) \cdot \vec{n}(x) &= -q_\mathcal{N}(x), && x \in \Gamma_\mathcal{N},\\
    \label{eq:pde-head-bc}
    u(x) &= u_\mathcal{D}(x), && x \in \Gamma_\mathcal{D},
\end{align}
where the prescribed flux $q_\mathcal{N} \in \mathbb{R}$ at the Neumann boundary $\Gamma_\mathcal{N}$ is in the outward direction $\vec{n}$ normal to $\Gamma_\mathcal{N}$, and $u_\mathcal{D} \in \mathbb{R}$ is the prescribed hydraulic head at the Dirichlet boundary $\Gamma_{\mathcal{D}}$.
We assume that all boundary conditions are known.

We evaluate the performance of the proposed BA-$K$D and BA-$K \times$1D methods by modeling the $u$ response at $N_u = 323$ observation locations as a function of the log-transmissivity field $y \coloneqq \log T$; that is, we set $\mathbf{g}(y) = \mathcal{G}(y)(X_u)$, where $X_u$ is the set of $N_u$ head observation locations.
We employ these BA-based low-dimensional surrogate models to address two computational tasks: (i) estimating the probability distribution of the observables via sampling for an uncertain $y$ field modeled as a GP, and (ii) estimating the $y$ field from noisy measurements of the observables.

In both tasks, we consider two parameterizations: The ``unconditional'' model is the KLE~\labelcref{eq:kle} $\tilde{y}(x; \bm{\xi}^{\text{unc}})$ corresponding to the unconditional GP with constant mean and Mat\'{e}rn-$5/2$ covariance kernel.
the ``conditional'' parameterization is the CKLE~\cref{eq:ckle-y} $\tilde{y}^c(\bm{x, \xi^{\text{cond}}})$ constructed by conditioning the unconditional GP on $N_y$ direct observations of the log-transmissivity field.
For both the unconditional and conditional models and for all values of $N_y$ considered, we set $N_{\xi} = 1,000$.
For the unconditional and conditional cases, the observable functions are, respectively, $\tilde{\mathbf{g}}(\bm{\xi}^{\text{unc}}) \coloneqq \mathbf{g}(\tilde{y}(x; \bm{\xi}^{\text{unc}}))$ and $\tilde{\mathbf{g}}^c(\bm{\xi}^{\text{cond}}) \coloneqq \mathbf{g}^c(\tilde{y}(x; \bm{\xi}^{\text{cond}}))$.

The experimental details including the finite volume discretization of the Hanford Site, the source and availability of the hydraulic head $u$ and log-transmissitivity $y$ reference fields, and the assumptions made for formulating the BVP~\cref{eq:pde,eq:pde-flux-bc,eq:pde-head-bc} can be found 
in~\cite{yeung-2022-wrr,yeung-2023-jcp}.
All BA-PCE surrogate models are constructed using a degree $p = 3$ Hermite polynomial expansion, and level $5$ Smolyak-Gauss-Hermite quadrature rules.
All algorithms are implemented in Python using the NumPy and SciPy packages.
All simulations are performed using a 3.2~GHz 8-core Intel Xeon W CPU and 32~GB of 2666~MHz DDR4 RAM\@.

\subsection{Accuracy of BA-based low-dimensional surrogates}\label{sec:results-accuracy}

We first evaluate the accuracy of the BA surrogate modeling algorithm (\cref{alg:ba}) and compare the performance of the surrogates for the unconditional and conditional cases.
Note that the observable functions for these cases are different in that they take different parameters; therefore, the performance of their corresponding surrogate models cannot be compared directly.
Rather, we compare their performance relative to the observable function each type of model (unconditional or conditional) aims to approximate.

To train the surrogate models, we assemble the training datasets\linebreak $\{ \bm{\Xi}^{\text{unc}}_{\text{train}}, \mathbf{U}^{\text{unc}}_{\text{train}} \}$ and $\{ \bm{\Xi}^{\text{cond}}_{\text{train}}, \mathbf{U}^{\text{cond}}_{\text{train}} \}$, each consisting of $q = 5,000$ random vectors $\bm{\xi}^{(i)}$ drawn form the standard multivariate normal distribution $p(\bm{\xi}) = \mathcal{N}(\mathbf{0}, \mathbf{I}_{N_{\xi}})$, together with the corresponding queries $\tilde{\mathbf{g}}(\bm{\xi}^{(i)})$ for the unconditional case and $\tilde{\mathbf{g}}^c(\bm{\xi}^{(i)})$ for the conditional case.
Additionally, we assemble testing datasets $\{ \bm{\Xi}^{\text{unc}}_{\text{test}}, \mathbf{U}^{\text{unc}}_{\text{test}} \}$ and $\{ \bm{\Xi}^{\text{cond}}_{\text{test}}, \mathbf{U}^{\text{cond}}_{\text{test}} \}$ in the same manner for $q = 5,000$ to be used to evaluate the out-of-sample performance of the surrogate models.

\begin{figure}
    \centering
    \includegraphics{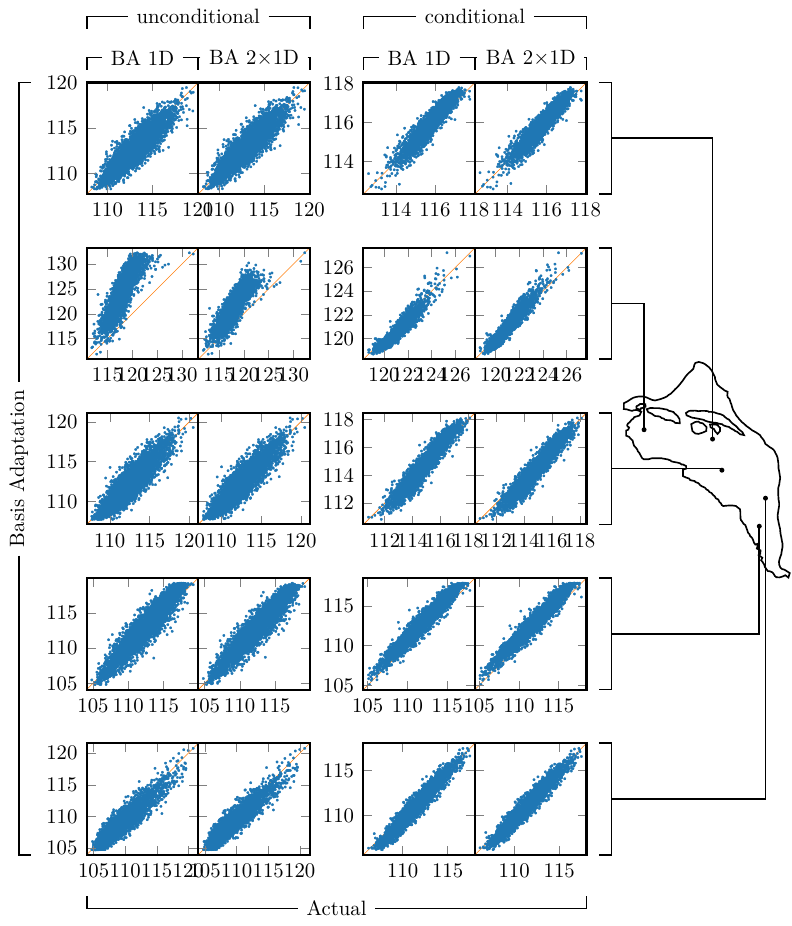}
    \caption{Scatter plots of BA-$1$D and BA-$2{\times}1$D estimates of hydraulic head compared to training data for $N_u = 5$ head observation locations.
      Results for the unconditional case are shown on the left column and for the conditional case on the right column.
    }\label{fig:scatter_plots_training}
\end{figure}

\begin{figure}
    \centering
    \includegraphics{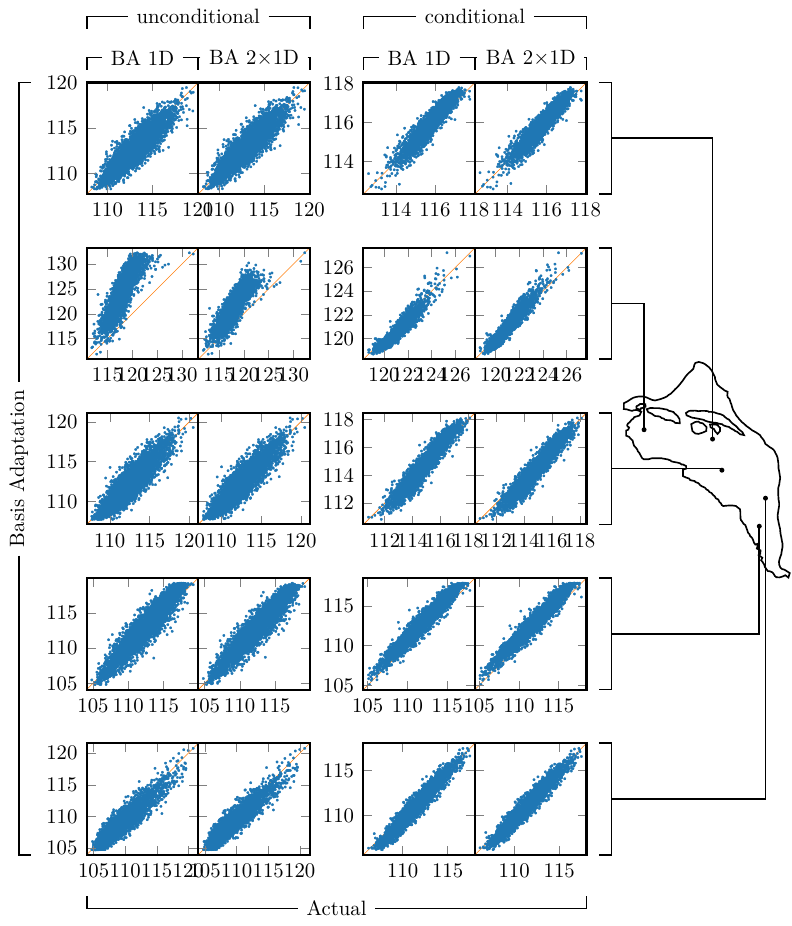}
    \caption{Scatter plots of BA-$1$D and BA-$2{\times}1$D estimates of hydraulic head compared to testing data for $N_u = 5$ head observation locations.
      Results for the unconditional case are shown on the left column and for the conditional case on the right column.
    }\label{fig:scatter_plots_testing}
\end{figure}

We apply~\cref{alg:ba} to train the surrogate models, which we use to estimate the observable function for each $\bm{\xi}^{(i)}$ in the training and testing datasets.
\Cref{fig:scatter_plots_training} shows scatter plots of the training data compared to the estimates of the BA-$1$D and BA-$2{\times}1$D algorithms for a subset of $5$ head observation locations from the set of $N_u = 323$ locations.
We chose $N_y = 100$ for the conditional case.
Similarly, \cref{fig:scatter_plots_testing} shows the scatter plots for the testing data.
We find that the use of the conditional CKLE-based parameterization significantly improves the accuracy of the BA-based low-dimensional surrogates of hydraulic head observables.
Specifically, while in the unconditional case the scatter plots of training data versus surrogate model predictions deviate substantially from the unit line, for the conditional case the scatter plots are all concentrated along the unit line, indicating good predictive performance.
We observe the same behavior in~\cref{fig:scatter_plots_testing} for the testing data, which indicates good out-of-sample predictive performance; nevertheless, we note that the spread of the scatter plots is wider for testing data compared to training data, which indicates that the surrogate models perform better for training data than for testing data.

\begin{table}[!htbp]
    \centering
    \caption{Root mean square errors of the BA-$1$D, BA-$2{\times}1$D, and BA-$2$D surrogate model predictions with respect to the training data for $N_u = 5$ head observation locations.
      The CKLE representation of the $y$ field is trained using $N_y = 25$, $50$, $100$, and $200$ log-transmissivity observation locations.
    }\label{tab:rmse}%
    \begin{tblr}{
        colspec = {*{2}{X[c]}*{5}{X[si={table-format=1.4}, c]}},
        row{1} = guard,
        hspan = minimal,
        hline{1,Z} = {1pt, solid},
        hline{2,5,8,11} = solid,
        cell{1}{3} = {c = 5}{c},
        cell{2,5,8,11}{1} = {r = 3}{c}
    }
        $N_y$ & BA & \begin{tikzpicture} [
            remember picture,
            baseline = (current bounding box.south),
            x = 10em,
            y = 10em,
            every circle/.style = { radius = 0.01 }
        ]
            \draw plot file {figures/hanford_boundary.txt};
            \fill (0.23125, 0.65625) coordinate (c1) circle;
            \fill (0.54375, 0.49375) coordinate (c2) circle;
            \fill (0.69375, 0.26875) coordinate (c3) circle;
            \fill (0.71875, 0.38125) coordinate (c4) circle;
            \fill (0.50625, 0.61875) coordinate (c5) circle;
            \foreach \c in {1,...,5}
                \draw (c\c) -| ($(0.5, 0) + ({(\c-3)*(\linewidth/5 + 2.5pt)}, 0)$) -- +(-1em, 0) -- +(1em, 0);
        \end{tikzpicture} & & & & \\
        25 & 1D & 0.8377 & 1.0022 & 0.9896 & 0.9730 & 0.9108 \\
        & 2$\times$1D & 0.8304 & 1.0052 & 0.9750 & 0.8829 & 0.9114 \\
        & 2D & 0.8224 & 1.0017 & 0.9772 & 0.9504 & 0.9108 \\
        50 & 1D & 0.5176 & 0.7387 & 0.7499 & 0.8221 & 0.6524 \\
        & 2$\times$1D & 0.5187 & 0.7316 & 0.7344 & 0.8719 & 0.6384 \\
        & 2D & 0.5108 & 0.7386 & 0.7343 & 0.8196 & 0.6506 \\
        100 & 1D & 0.3578 & 0.4179 & 0.6793 & 0.6506 & 0.3047 \\
        & 2$\times$1D & 0.3418 & 0.4151 & 0.6744 & 0.6149 & 0.2921 \\
        & 2D & 0.3416 & 0.4174 & 0.6701 & 0.6475 & 0.3013 \\
        200 & 1D & 0.2255 & 0.1333 & 0.3069 & 0.2325 & 0.1026 \\
        & 2$\times$1D & 0.2160 & 0.1311 & 0.3022 & 0.2338 & 0.1006 \\
        & 2D & 0.2131 & 0.1332 & 0.3066 & 0.2320 & 0.1020 \\
    \end{tblr}
\end{table}

\begin{table}[!htbp]
    \centering
    \caption{Root mean square errors of the BA-$1$D, BA-$2{\times}1$D, and BA-$2$D surrogate model predictions with respect to the testing data for $N_u = 5$ head observation locations.
      The CKLE representation of the $y$ field is trained using $N_y = 25$, $50$, $100$, and $200$ log-transmissivity observation locations.
    }\label{tab:rmse_testing}%
    \begin{tblr}{
        colspec = {*{2}{X[c]}*{5}{X[si={table-format=1.4}, c]}},
        row{1} = guard,
        hspan = minimal,
        hline{1,Z} = {1pt, solid},
        hline{2,5,8,11} = solid,
        cell{1}{3} = {c = 5}{c},
        cell{2,5,8,11}{1} = {r = 3}{c}
    }
        $N_y$ & BA & \begin{tikzpicture} [
            remember picture,
            baseline = (current bounding box.south),
            x = 10em,
            y = 10em,
            every circle/.style = { radius = 0.01 }
        ]
            \draw plot file {figures/hanford_boundary.txt};
            \fill (0.23125, 0.65625) coordinate (c1) circle;
            \fill (0.54375, 0.49375) coordinate (c2) circle;
            \fill (0.69375, 0.26875) coordinate (c3) circle;
            \fill (0.71875, 0.38125) coordinate (c4) circle;
            \fill (0.50625, 0.61875) coordinate (c5) circle;
            \foreach \c in {1,...,5}
                \draw (c\c) -| ($(0.5, 0) + ({(\c-3)*(\linewidth/5 + 2.5pt)}, 0)$) -- +(-1em, 0) -- +(1em, 0);
        \end{tikzpicture} & & & & \\
        25 & 1D & 0.8374 & 1.0020 & 0.9896 & 0.9731 & 0.9106 \\
        & 2$\times$1D & 0.8298 & 1.0050 & 0.9749 & 0.8832 & 0.9113 \\
        & 2D & 0.8222 & 1.0015 & 0.9772 & 0.9505 & 0.9106 \\
        50 & 1D & 0.5161 & 0.7390 & 0.7500 & 0.8231 & 0.6533 \\
        & 2$\times$1D & 0.5177 & 0.7321 & 0.7342 & 0.8703 & 0.6396 \\
        & 2D & 0.5093 & 0.7389 & 0.7341 & 0.8207 & 0.6515 \\
        100 & 1D & 0.3574 & 0.4185 & 0.6804 & 0.6511 & 0.3050 \\
        & 2$\times$1D & 0.3417 & 0.4155 & 0.6749 & 0.6151 & 0.2923 \\
        & 2D & 0.3414 & 0.4180 & 0.6714 & 0.6479 & 0.3015 \\
        200 & 1D & 0.2259 & 0.1326 & 0.3056 & 0.2327 & 0.1026 \\
        & 2$\times$1D & 0.2160 & 0.1303 & 0.3017 & 0.2334 & 0.1005 \\
        & 2D & 0.2134 & 0.1325 & 0.3053 & 0.2322 & 0.1019 \\
    \end{tblr}
\end{table}

\Cref{fig:scatter_plots_training,fig:scatter_plots_testing} do not show an appreciable difference between the predictive performance of the BA-$1$D and BA-$2{\times}1$D models.
To further verify this, in~\cref{tab:rmse} we present the root mean square error (RMSE) of the surrogate model estimates with respect to the training data for the conditional case.
The $2$D models perform on average about $1$\% better than the $1$D model, although whether the $2$D or the $2{\times}1$D model is the better performing depends on the head observation location.
\cref{tab:rmse_testing} demonstrates similar behavior with the testing data.
Nevertheless, \cref{sec:results-ba-map} shows that identifying a second dominant direction of variation improves the performance of the surrogate models for parameter estimation.

Overall, the results shown in~\cref{fig:scatter_plots_training,fig:scatter_plots_testing} and~\cref{tab:rmse} indicate that, while the variation of the observable response is mostly concentrated along the linear subspace $\operatorname{span} \{ \hat{\mathbf{a}}_1 \}$, it is not entirely resolved by a 1D ridge function model, and that 2D ridge function surrogate models modestly improve accuracy.
Other possible sources, such as the assumption of a linear transformation from parameter space to the effective coordinates, will be studied in future work.

\subsection{Uncertainty quantification}\label{sec:results-uq}

We first evaluate the performance of the proposed surrogate modeling algorithm by applying these surrogate models to uncertainty quantification tasks.
Specifically, we use sampling to estimate the marginal PDF of each component of the observation functions $\tilde{\mathbf{g}}(\bm{\xi}^{\text{unc}})$ and $\tilde{\mathbf{g}}^c(\bm{\xi}^{\text{cond}})$ when $\bm{\xi}^{\text{unc}}$ and $\bm{\xi}^{\text{cond}}$ are random with distribution $p(\bm{\xi}^{\text{unc}}) = p(\bm{\xi}^{\text{cond}}) = \mathcal{N}(\mathbf{0}, \mathbf{I}_{N_{\xi}})$.
For this, we query the surrogate model for each of the samples in the testing dataset (noting that the data-generating distribution is equal to the distribution we are imposing here for $\bm{\xi}^{\text{unc}}$ and $\bm{\xi}^{\text{cond}}$).
We then estimate the PDF from these samples via kernel density estimation (KDE) with a Gaussian kernel.
The kernel bandwidth is estimated using the algorithm presented in~\cite{scott-multivariate-2015}.
Finally, we estimate the actual PDFs via KDE using the testing dataset samples and compare the testing data PDF and the surrogate model-generated PDF.

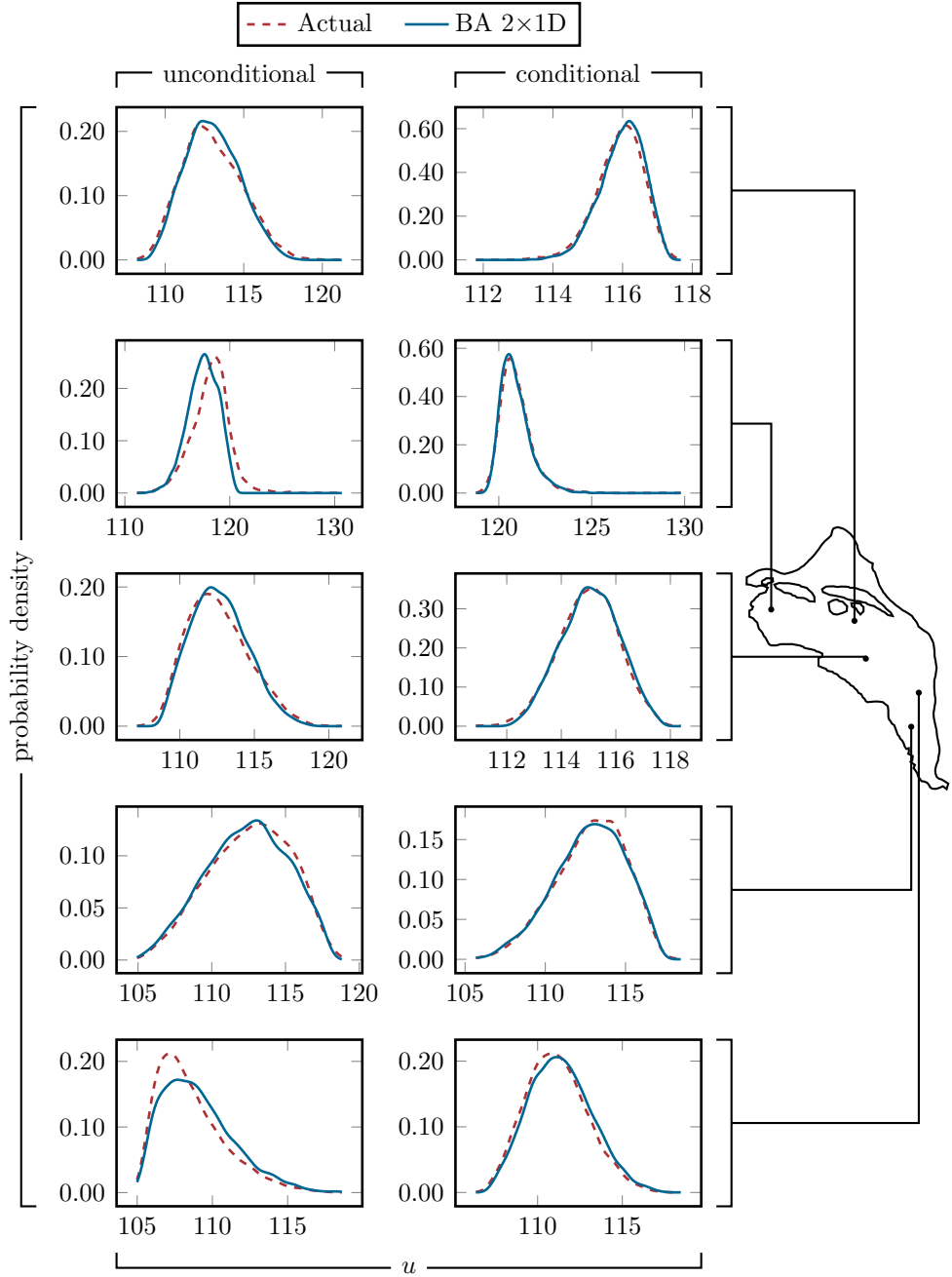
\begin{figure}[!htbp]
    \centering
    \begin{tikzpicture}
        \begin{groupplot}[
            group style = {
                group name = plots,
                group size = 2 by 5,
                vertical sep = 5ex,
                horizontal sep = 7ex,
            },
            pdf plots
        ]
            \pgfplotstableread{figures/u_ba_vs_mc_pdf_NY=100_Nu=323_NYxi=1000_Nens=5000_1d_dim=2_deg=3_n_gauss=5_sigma=1e-12_uncond_testing_RF1_1x.txt}{\pdfuncond}
            \pgfplotstableread{figures/u_ba_vs_mc_pdf_NY=100_Nu=323_NYxi=1000_Nens=5000_1d_dim=2_deg=3_n_gauss=5_sigma=1e-12_cond_testing_RF1_1x.txt}{\pdfcond}
            \def\myPlots{}
            \pgfplotsforeachungrouped \i in {942,423,321,124,555} {
                \pgfplotsforeachungrouped \c in {0,1} {
                    \eappto\myPlots{\noexpand\nextgroupplot [legend to name = pdftestinglegend\i\c]}
                    \pgfplotsforeachungrouped \y in {mc, ba_2x1d} {
                        \eappto\myPlots{\noexpand\addplot+ table [x=\i_x,y=\i_\y] \ifnum\c=0{\noexpand\pdfuncond}\else{\noexpand\pdfcond}\fi;}
                    }
                }
            }
            \myPlots
        \end{groupplot}
        
        \node [left = of plots c1r3, anchor = south, rotate = 90, align = center] (ylabel) {probability density};
        \node [below = 1.5em] at ($(plots c1r5.south)!0.5!(plots c2r5.south)$) (xlabel) {$u$};
        \node [above = 0.5em of plots c1r1.north, anchor = south] (uncond) {unconditional};
        \node [above = 0.5em of plots c2r1.north, anchor = south] (cond) {conditional};
        \node [above = 0.5em, anchor = south] at ($(uncond)!0.5!(cond)$) {\ref*{pdftestinglegend1240}};
        \draw (uncond) -- (uncond -| plots c1r1.east) -- +(0, -0.5em)
            (uncond) -- (uncond -| plots c1r1.west) -- +(0, -0.5em)
            (cond) -- (cond -| plots c2r1.east) -- +(0, -0.5em)
            (cond) -- (cond -| plots c2r1.west) -- +(0, -0.5em)
            (xlabel) -- (xlabel -| plots c2r5.east) -- +(0, 0.5em)
            (xlabel) -- (xlabel -| plots c1r5.west) -- +(0, 0.5em)
            (ylabel) -- (ylabel |- plots c1r1.north) -- +(0.5em, 0)
            (ylabel) -- (ylabel |- plots c1r5.south) -- +(0.5em, 0);
        \begin{scope} [
            shift = {($(plots c2r3.east) + (0, -5em)$)},
            x = 10em,
            y = 10em,
            every circle/.style = { radius = 0.01 }
        ]
            \draw plot file {figures/hanford_boundary.txt};
            \fill (0.50625, 0.61875) coordinate (c1) circle;
            \fill (0.23125, 0.65625) coordinate (c2) circle;
            \fill (0.54375, 0.49375) coordinate (c3) circle;
            \fill (0.69375, 0.26875) coordinate (c4) circle;
            \fill (0.71875, 0.38125) coordinate (c5) circle;
        \end{scope}
        \foreach \r in {1,...,5} {
            \draw ([xshift = 0.5em] plots c2r\r.north east) -- ++(0.5em, 0) -- coordinate (index\r)
                ([xshift = 1em] plots c2r\r.south east) -- ++(-0.5em, 0)
                (index\r) -| (c\r);
        }
    \end{tikzpicture}
    \caption{PDFs of the observable function corresponding to $N_u = 5$ head observation locations for the testing dataset at various locations}\label{fig:pdf_plots_testing}
\end{figure}

\Cref{fig:pdf_plots_testing} shows the PDFs estimated via BA surrogate modeling compared to the PDFs estimated from the testing data.
Given that the differences in RMSE between the BA-$1$D, BA-$2{\times}1$D, and BA-$2$D models are small, here we only show results for the BA-$2$D model.
The surrogate model PDF estimates are more accurate for the conditional case than for the unconditional case.
Furthermore, consistent with~\cite{tipireddy-2020-jcp}, the PDFs for the conditional case are narrower (as seen by the narrower range of values in the horizontal axis), which indicates that by conditioning the $y$ field model in available direct observations we can appreciably reduce the uncertainty in quantities of interest.

\begin{table}[!htbp]
    \centering
    \caption{Kullback–Leibler divergence of the basis adaptation approximations vs.\@ Monte Carlo targets as functions of $N_{\mathbf{y}_{\mathrm{s}}}$ at 5 cell locations, when evaluated using $\mathcal{X}_{\text{testing}}$.}\label{tab:kl_divergence}%
    \begin{tblr}{
        colspec = {*{2}{X[c]}*{5}{X[si={table-format=1.4}, c]}},
        row{1} = guard,
        hspan = minimal,
        hline{1,Z} = {1pt, solid},
        hline{2,5,8,11} = solid,
        cell{1}{3} = {c = 5}{c},
        cell{2,5,8,11}{1} = {r = 3}{c}
    }
        $N_y$ & BA & \begin{tikzpicture} [
            remember picture,
            baseline = (current bounding box.south),
            x = 10em,
            y = 10em,
            every circle/.style = { radius = 0.01 }
        ]
            \draw plot file {figures/hanford_boundary.txt};
            \fill (0.23125, 0.65625) coordinate (c1) circle;
            \fill (0.54375, 0.49375) coordinate (c2) circle;
            \fill (0.69375, 0.26875) coordinate (c3) circle;
            \fill (0.71875, 0.38125) coordinate (c4) circle;
            \fill (0.50625, 0.61875) coordinate (c5) circle;
            \foreach \c in {1,...,5}
                \draw (c\c) -| ($(0.5, 0) + ({(\c-3)*(\linewidth/5 + 2.5pt)}, 0)$) -- +(-1em, 0) -- +(1em, 0);
        \end{tikzpicture} & & & & \\
        25 & 1D & 0.0373 & 0.0808 & 0.0727 & 0.0680 & -0.0230 \\
        & 2$\times$1D & 0.0700 & 0.0623 & 0.0423 & -0.0295 & 0.0679 \\
        & 2D & 0.0223 & 0.0362 & 0.0717 & 0.1038 & 0.0204 \\
        50 & 1D & 0.0034 & 0.0097 & 0.0122 & 0.0323 & 0.0632 \\
        & 2$\times$1D & 0.0030 & 0.0414 & -0.0265 & -0.0677 & 0.0265 \\
        & 2D & 0.0088 & 0.0044 & 0.0717 & 0.0048 & -0.0166 \\
        100 & 1D & -0.0349 & 0.0248 & 0.0388 & -0.0069 & -0.0182 \\
        & 2$\times$1D & -0.0018 & -0.0044 & 0.0560 & -0.0249 & 0.0121 \\
        & 2D & -0.0074 & 0.0239 & 0.0296 & 0.0214 & 0.0049 \\
        200 & 1D & -0.0141 & -0.0191 & -0.0303 & -0.0015 & 0.0078 \\
        & 2$\times$1D & -0.0281 & 0.0171 & 0.0468 & 0.0221 & -0.0098 \\
        & 2D & -0.0016 & -0.0003 & 0.0365 & 0.0201 & 0.0573 \\
    \end{tblr}
\end{table}

Additionally, we compute the Kullback–Leibler divergence between the surrogate model PDF estimates and the PDFs estimated from the testing dataset for the BA-$1$D, BA-$2{\times}1$D, and BA-$2$D models and for the same $N_u = 5$ head observation locations shown in~\cref{fig:pdf_plots_testing}.
We perform these calculations for various values of $N_y$.
The results are shown in~\Cref{tab:kl_divergence}.
We observe mixed results using this metric when comparing the three BA surrogate models.
Compared to~\Cref{tab:rmse}, we do not find the $2$D models to perform better than the $1$D model; in fact, all BA surrogate models have comparable performance.
While it may seem counterintutive to find differences in point-wise (in parameter space) but no differences in the resulting PDFs, one can think of the joint distribution of the exact observables and the observables predicted by the surrogate models (visualized by the scatter plots of samples shown in~\cref{fig:scatter_plots_training,fig:scatter_plots_testing}), which may have equal marginal distributions even though the components are not perfectly correlated.

\subsection{BA-MAP for parameter estimation}\label{sec:results-ba-map}

In the following, we refer to the inverse problem formulation~\labelcref{eq:inverse-problem} with BA surrogate models as BA-MAP.
We solve the corresponding nonlinear least-squares problem using SciPy's implementation of the trust region reflective algorithm~\cite{branch-subspace-1999}.
We compare the estimates of log-transmissivity obtained with BA-MAP against the CKLEMAP algorithm~\cite{yeung-2023-jcp}.
In CKLEMAP, we also use a CKLE representation of the parameter field, but solve the BVP~\cref{eq:pde,eq:pde-flux-bc,eq:pde-head-bc} to evaluate the hydraulic head field and the data misfit term for a given $\bm{\xi}^{\text{cond}}$.

\begin{figure}[!htbp]
    \centering
    \begin{tblr}{
        colspec = {X[1.1,c,m]*{4}{X[c,m]}l},
        hline{1,Z} = {1pt, solid},
        hline{2} = solid,
        cell{1}{2} = {c = 4}{c}
    }
        reference & \adjustimage{scale=0.25}{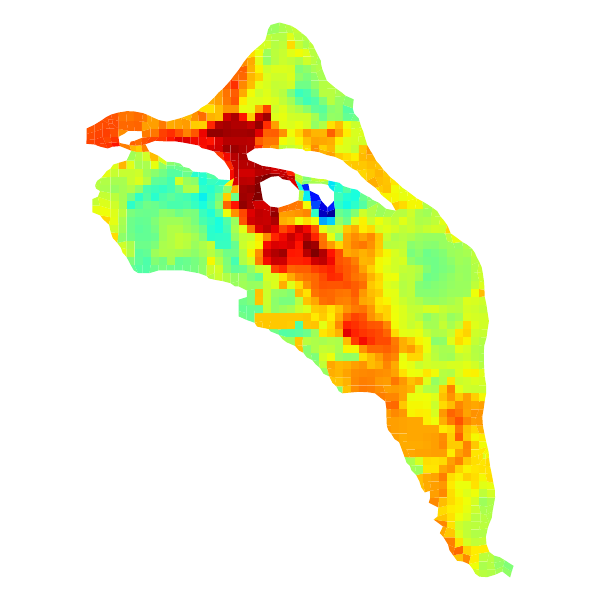} &
        \colorbar{0}{12}{{0,2,4,...,12}} \\
        $N_y$ & 25 & 50 & 100 & 200 & \\
        observation locations &
        \adjustimage{scale=0.25}{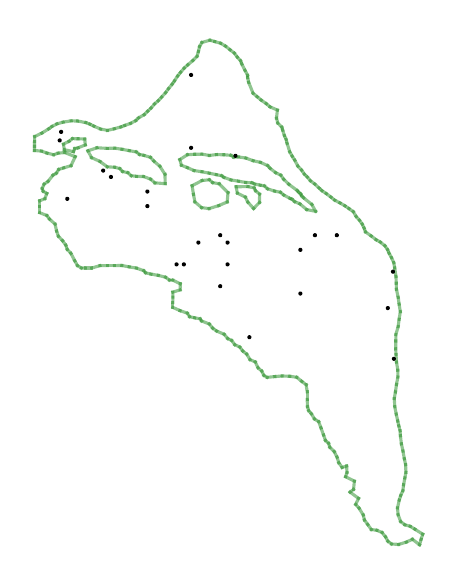} &
        \adjustimage{scale=0.25}{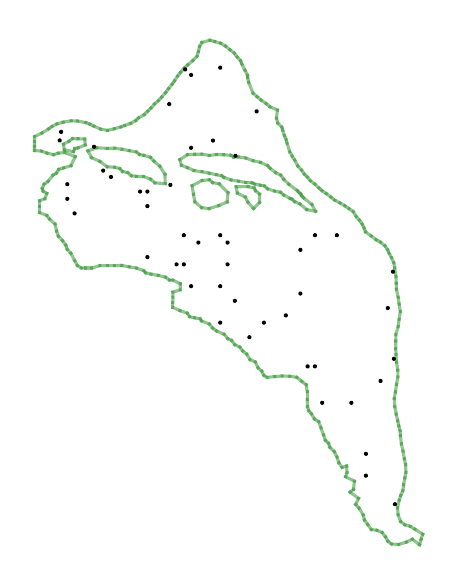} &
        \adjustimage{scale=0.25}{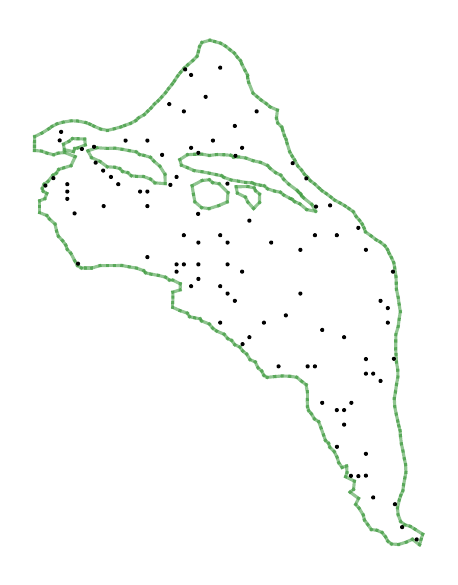} &
        \adjustimage{scale=0.25}{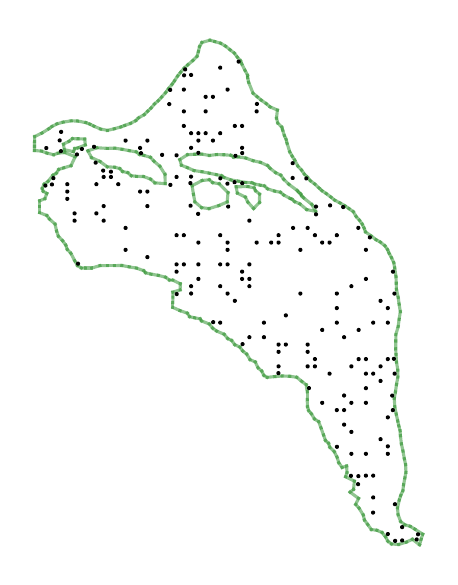} & \\
        CKLEMAP estimates &
        \adjustimage{scale=0.25}{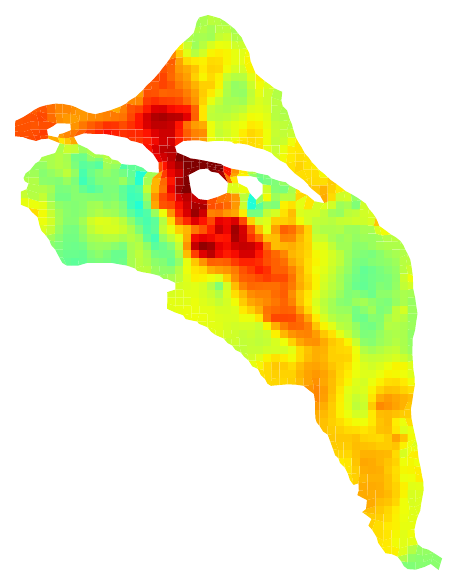} &
        \adjustimage{scale=0.25}{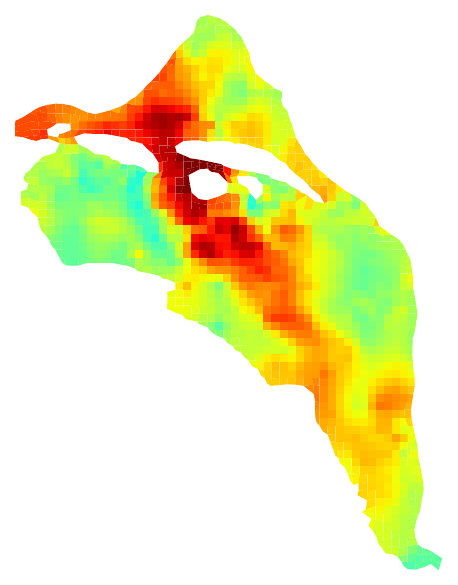} &
        \adjustimage{scale=0.25}{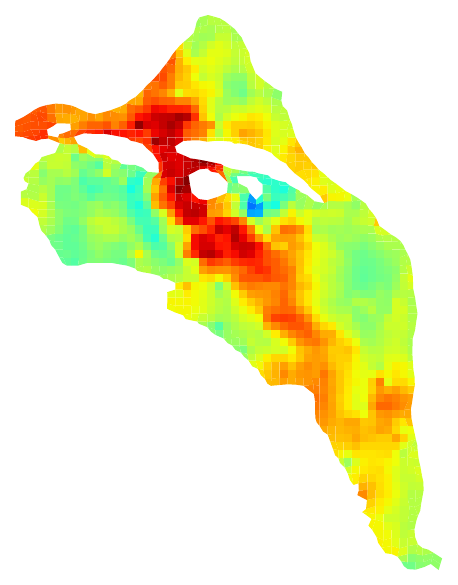} &
        \adjustimage{scale=0.25}{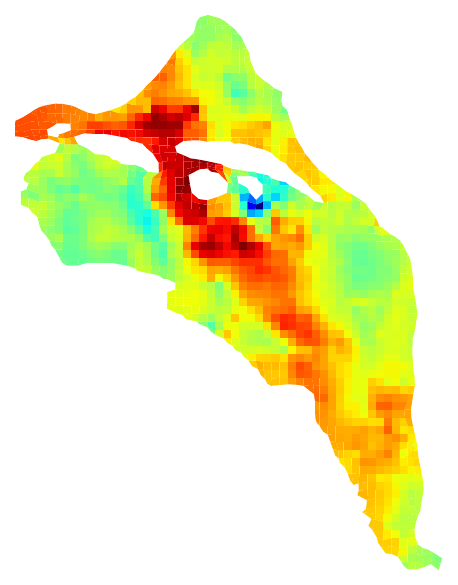} &
        \colorbar{0}{12}{{0,2,4,...,12}} \\
        CKLEMAP point errors &
        \adjustimage{scale=0.25}{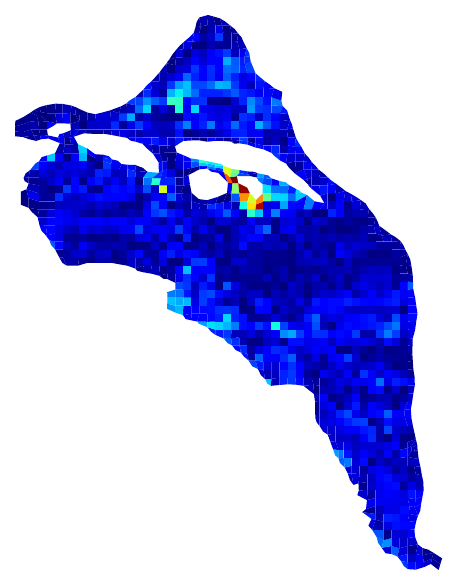} &
        \adjustimage{scale=0.25}{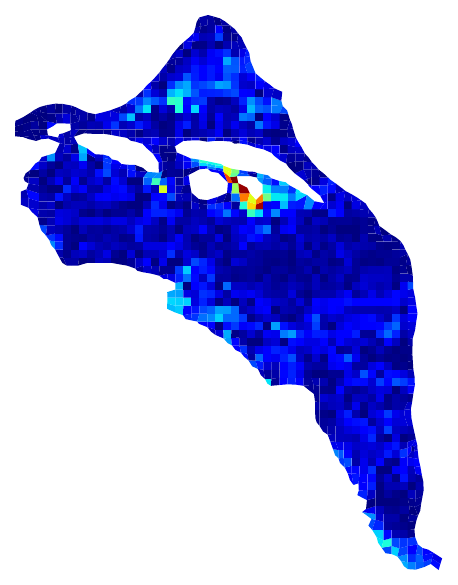} &
        \adjustimage{scale=0.25}{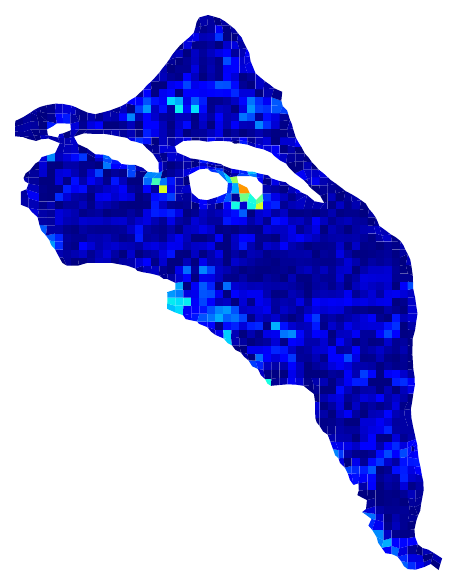} &
        \adjustimage{scale=0.25}{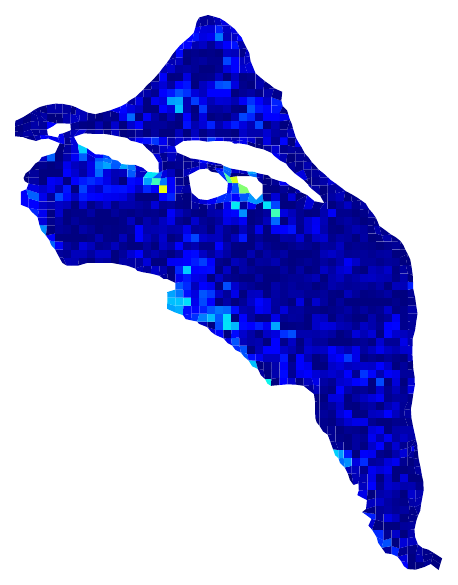} &
        \colorbar{0}{6}{{0,1,2,...,6}} \\
        BA-MAP estimates &
        \adjustimage{scale=0.25}{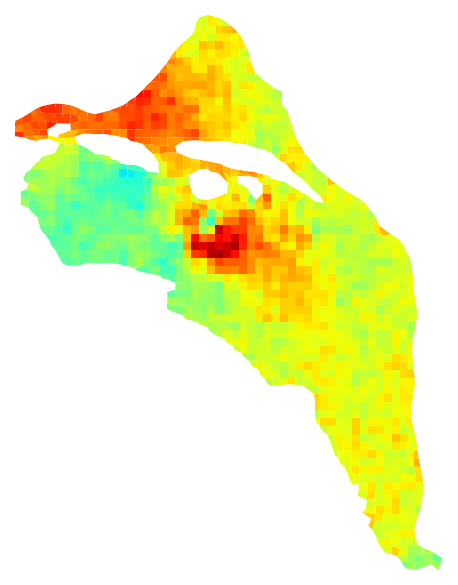} &
        \adjustimage{scale=0.25}{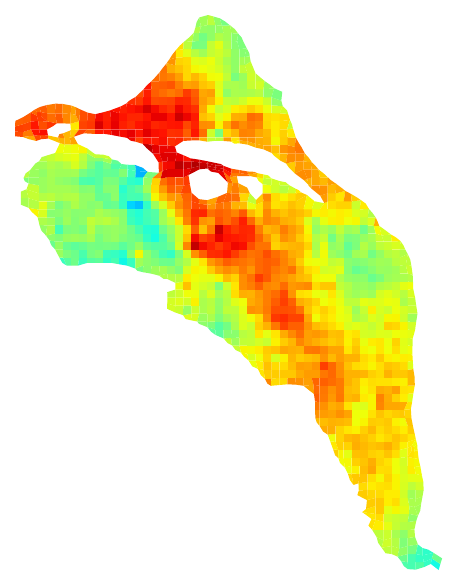} &
        \adjustimage{scale=0.25}{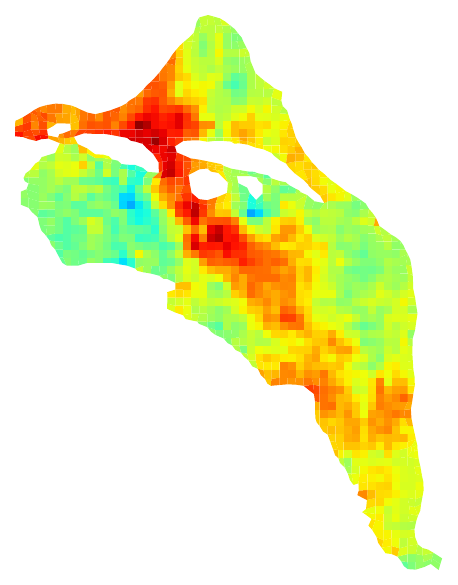} &
        \adjustimage{scale=0.25}{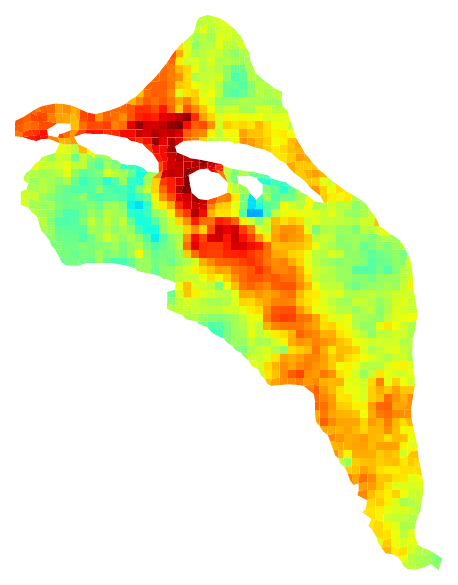} &
        \colorbar{0}{12}{{0,2,4,...,12}} \\
        BA-MAP point errors &
        \adjustimage{scale=0.25}{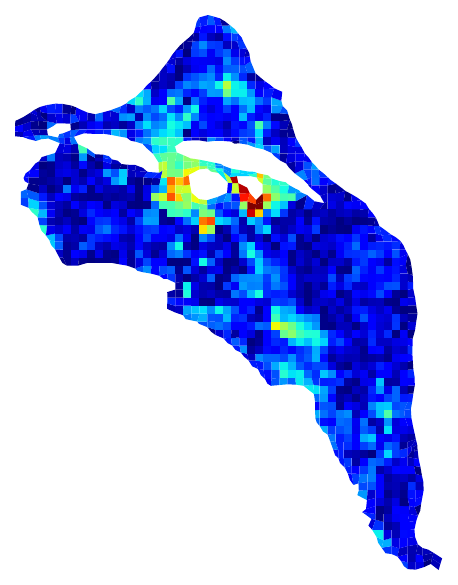} &
        \adjustimage{scale=0.25}{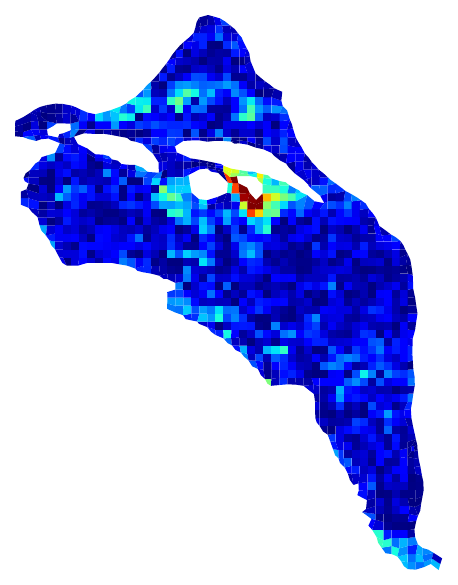} &
        \adjustimage{scale=0.25}{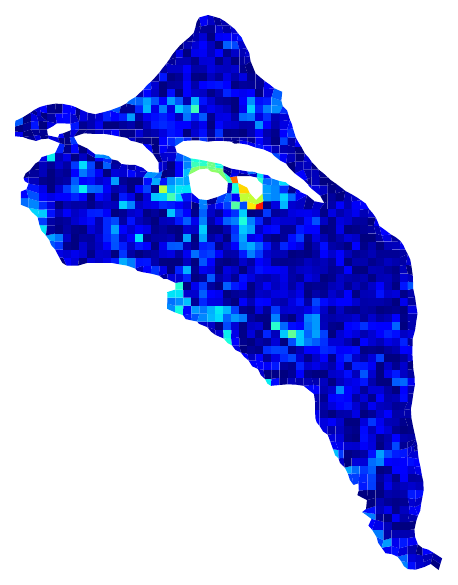} &
        \adjustimage{scale=0.25}{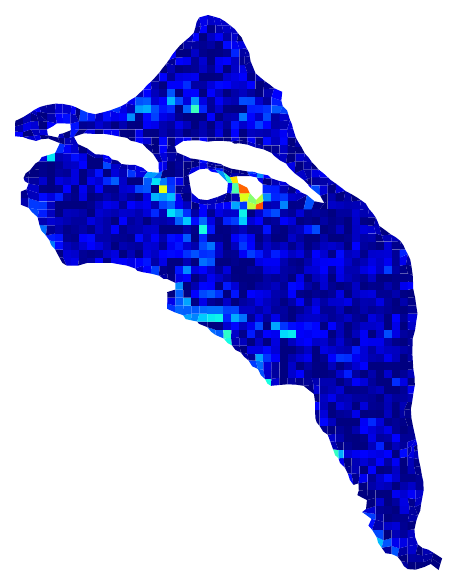} &
        \colorbar{0}{6}{{0,1,2,...,6}} \\
    \end{tblr}
    \caption{Reference log-transmissivity field $y$ fields, CKLEMAP and BA-MAP estimates and point errors for $N_y = 25$, $50$, $100$, and $200$.}\label{fig:hanford_1x}
\end{figure}

\Cref{fig:hanford_1x} shows the CKLEMAP and CKLE-based BA-MAP (BA-$2{\times}1$D)
estimates computed for various values of $N_y$,
as well as the log-transmissivity observation locations and the absolute point errors.
Although the accuracy of basis adaptation estimates are not as good as CKLEMAP, it provides a good estimate in most cases except for the case $N_y = 25$.
We present in~\Cref{tab:uncond_ba_map_results,tab:cond_ba_map_results} the number of least-square iterations, the relative $\ell_2$ errors, and the absolute $\ell_\infty$ errors of CKLEMAP and BA-MAP with BA-$1$D and BA-$2{\times}1$D surrogates, both for the unconditional and conditional cases.
We chose $\gamma = 10^{-6}$ for the conditional case.
For the unconditional case, we find that we must use the larger value $\gamma = 10^{-1}$.
We can observe that using the $2$D surrogate model modestly reduces the relative $\ell_2$ error.
Most importantly, the $2$D surrogate model drastically reduces the number of iterations of the optimization algorithm.
Furthermore, with the exception of the case $N_y = 25$, inversion with CKLE-based BA surrogate models leads to more accurate estimates than using BA surrogate models for the unconditional parameterization.
Finally, in general, BA-MAP is faster than CKLEMAP to compute per iteration (disregarding the training cost).

\begin{table}[!htbp]
    \centering
    \caption{Number of least-squares iterations, relative $\ell_2$ errors, and absolute $\ell_{\infty}$ errors of BA-MAP for the unconditional parameterization, computed for $N_y = 25$, $50$, $100$, and $200$.}\label{tab:uncond_ba_map_results}%
    \begin{tblr}{
        colspec = {X[1.5,l]X[2.5,l]*{4}{X[c]}},
        row{2} = c,
        vspan = even,
        hline{1,Z} = {1pt, solid},
        hline{3,5,7} = solid,
        cell{1}{3} = {c = 4}{c},
        cell{3,5,7}{1} = {r = 2}{l}
    }
        & & $N_y$ & & & \\
        & solver & 25 & 50 & 100 & 200 \\
        {least-square\\iterations} & BA-MAP (1D) & 36 & 37 & 96 & 282 \\
        & BA-MAP ($2{\times}1$D) & 78 & 46 & 80 & 296 \\
        {relative\\$\ell_2$ error} & BA-MAP (1D) & 0.159 & 0.145 & 0.126 & 0.112 \\
        & BA-MAP ($2{\times}1$D) & 0.151 & 0.141 & 0.124 & 0.111 \\
        {absolute\\$\ell_\infty$ error} & BA-MAP (1D) & 7.47 & 7.46 & 5.49 & 5.95 \\
        & BA-MAP ($2{\times}1$D) & 8.14 & 7.59 & 5.45 & 6.31 \\
    \end{tblr}
\end{table}

\begin{table}[!htbp]
    \centering
    \caption{
      Number of least-squares iterations, relative $\ell_2$ errors, and absolute $\ell_{\infty}$ errors of CKLEMAP and BA-MAP for the CKLE-based parameterization, computed for $N_y = 25$, $50$, $100$, and $200$.}\label{tab:cond_ba_map_results}%
    \begin{tblr}{
        colspec = {X[1.5,l]X[2.5,l]*{4}{X[c]}},
        row{2} = c,
        hline{1,Z} = {1pt, solid},
        hline{3,6,9} = solid,
        cell{1}{3} = {c = 4}{c},
        cell{3,6,9}{1} = {r = 3}{l}
    }
        & & $N_y$ & & & \\
        & solver & 25 & 50 & 100 & 200 \\
        {least-square\\iterations} & CKLEMAP & 63 & 41 & 44 & 25 \\
        & BA-MAP (1D) & 22045 & 24429 & 10 & 2538 \\
        & BA-MAP ($2{\times}1$D) & 72 & 42 & 23 & 65 \\
        {relative\\$\ell_2$ error} & CKLEMAP & 0.102 & 0.101 & 0.075 & 0.067 \\
        & BA-MAP (1D) & 0.169 & 0.139 & 0.113 & 0.081 \\
        & BA-MAP ($2{\times}1$D) & 0.135 & 0.135 & 0.106 & 0.082 \\
        {absolute\\$\ell_\infty$ error} & CKLEMAP & 6.06 & 6.11 & 4.52 & 4.69 \\
        & BA-MAP (1D) & 8.39 & 8.36 & 5.10 & 4.79 \\
        & BA-MAP ($2{\times}1$D) & 8.34 & 8.34 & 5.28 & 4.90 \\
    \end{tblr}
\end{table}


%% file: conclusions.tex
In this work, we have studied the data-driven construction of low-\linebreak dimensional surrogate models for observables of physical systems governed by BVPs with heterogeneous parameter fields and their use for solving uncertainty quantification and parameter estimation problems.
We assume that the input parameter fields are represented using truncated KLEs, which define a finite-dimensional parameter space in terms of the KLE coefficients.
We construct these low-dimensional surrogate models as the composition of a linear transformation from the parameter space to a low-dimensional set of effective coordinates, and a PCE model mapping from the effective coordinates to observables.
To identify the linear rotation to effective coordinates, we use the BA method, which in its original formulation provides a one-dimensional transformation per scalar observable.
Additionally, we propose a heuristic for identifying additional effective coordinates for a scalar observable by regressing the residuals of (i.e., the difference between) training data and the one-dimensional model's predictions.

We consider low-dimensional surrogates of the hydraulic head response at observation wells for a two-dimensional confined aquifer model of the Hanford Site.
We find that by conditioning the KLE representation of the input log-transmissivity field on spatially sparse direct measurements of log-transmissivity, we can improve the accuracy of the low-dimensional surrogate models of the hydraulic head observations.
We also find that low-dimensional surrogate models based on these conditional representations lead to more accurate solutions to uncertainty quantification and parameter estimation problems than surrogate models based on unconditional representations.

For the hydraulic head response, we find that one-dimensional surrogate models are only slightly less acccurate overall than two-dimensional models, although the two-dimensional models perform better than one-dimensional models in parameter estimation tasks.
We hypothesize that this is because effective coordinates for these observables can only be approximated by linear transformations, and that nonlinear methods are necessary to more accurately identify effective coordinates.
In future work, we will explore the use of nonlinear transformations to effective coordinates for formulating low-dimensional surrogate models.
